\documentclass[11pt,a4paper]{article}
\usepackage[hyperref]{latell}
\usepackage{times}
\usepackage{latexsym}

\usepackage{microtype}

\usepackage{graphicx}
\usepackage{tabularray}
\usepackage{array}
\usepackage{float}
\usepackage{amssymb}
\usepackage{booktabs}
\usepackage{subcaption}
\usepackage[dvipsnames]{xcolor}
\usepackage{multirow}
\usepackage[table]{xcolor}   
\usepackage{siunitx}

\usepackage[shortlabels]{enumitem} %

\usepackage{ifthen}
\usepackage{amssymb}
\newboolean{showcomments}
\setboolean{showcomments}{true}
\ifthenelse{\boolean{showcomments}}
 { \newcommand{\mynote}[2]{
      \fbox{\bfseries\sffamily\scriptsize#1}
        {\small$\blacktriangleright$\textsf{\emph{#2}}$\blacktriangleleft$}}}
        { \newcommand{\mynote}[2]{}}

\usepackage[T1]{fontenc}

\usepackage[utf8]{inputenc}
\usepackage{tcolorbox}
\usepackage{listings}
\usepackage[dvipsnames]{xcolor}
\tcbuselibrary{listings,breakable}
\definecolor{darkgreen}{rgb}{0.0, 0.5, 0.0}
\definecolor{darkred}{rgb}{0.7, 0.0, 0.0}

\lstdefinelanguage{json}{
    basicstyle=\footnotesize\ttfamily,
    showstringspaces=false,
    breaklines=true,
    breakatwhitespace=true,
    literate=
     *{0}{{{\color{blue}0}}}{1}
      {1}{{{\color{blue}1}}}{1}
      {2}{{{\color{blue}2}}}{1}
      {3}{{{\color{blue}3}}}{1}
      {4}{{{\color{blue}4}}}{1}
      {5}{{{\color{blue}5}}}{1}
      {6}{{{\color{blue}6}}}{1}
      {7}{{{\color{blue}7}}}{1}
      {8}{{{\color{blue}8}}}{1}
      {9}{{{\color{blue}9}}}{1}
      {:}{{{\color{red}{:}}}}{1}
      {,}{{{\color{red}{,}}}}{1}
      {\{}{{{\color{orange}{\{}}}}{1}
      {\}}{{{\color{orange}{\}}}}}{1}
      {[}{{{\color{orange}{[}}}}{1}
      {]}{{{\color{orange}{]}}}}{1},
    string=[s]{"}{"},
}

\aclfinalcopy 
\def\aclpaperid{53} 

\title{LuxIT: A Luxembourgish Instruction Tuning Dataset from Monolingual Seed Data}

\author{Julian Valline, \space Cedric Lothritz, \space Siwen Guo, \space Jordi Cabot \\
  Luxembourg Institute of Science and Technology \\
  5 Av. des Hauts-Fourneaux \\
  L-4362 Esch-sur-Alzette \\
  \texttt{\{julian.valline, cedric.lothritz,} \\
  \texttt{siwen.guo, jordi.cabot\}@list.lu} }

\date{}

\begin{document}
\maketitle

\begin{abstract}
The effectiveness of instruction-tuned Large Language Models (LLMs) is often limited in low-resource linguistic settings due to a lack of high-quality training data. We introduce  LuxIT, a monolingual instruction tuning dataset for Luxembourgish developed to mitigate this challenge. We synthesize the dataset from a corpus of native Luxembourgish texts, utilizing DeepSeek-R1-0528, chosen for its shown proficiency in Luxembourgish. Following generation, we apply a quality assurance process, employing an LLM-as-a-judge approach,  retaining 227,507 high-quality instruction-answer pairs. To investigate the practical utility of the dataset, we fine-tune 14 smaller-scale LLMs ($\leq$15B parameters) on LuxIT and evaluate them on standardized Luxembourgish proficiency exams and five downstream NLP tasks. Training on LuxIT yields a mean accuracy change of +5.37 percentage points on language exams across all 14 models, with 12 of 14 showing improvement. On NLP downstream tasks, 9 of 14 models improve in macro-averaged F1, though gains on the two benchmarks do not systematically correlate. These results underscore the feasibility of leveraging monolingual synthetic data to improve LLM capabilities in low-resource languages, while highlighting the multi-faceted nature of language proficiency.

\end{abstract}
\section{Introduction}

In recent years, Large Language Models (LLMs) have demonstrated remarkable proficiency across a diverse range of natural language tasks \citep{zhang2025instructiontuninglargelanguage, NEURIPS2020_1457c0d6, touvron2023llama2openfoundation, NEURIPS2023_548a41b9, peng2023instructiontuninggpt4, li2023mimicitmultimodalincontextinstruction}. Their rapid and widespread adoption as virtual assistants \citep{NEURIPS2023_c2a8060f} has been largely driven by their accessibility through intuitive chat-based interfaces \citep{weber-etal-2024-investigating, touvron2023llama2openfoundation}. The training of these assistants involves multiple steps, beginning with a pre-training stage where the model learns from a vast text corpus using a self-supervised objective such as next-word prediction \citep{weber-etal-2024-investigating, touvron2023llama2openfoundation, NEURIPS2023_ac662d74}. The subsequent, critical step is
instruction tuning (IT) \citep{wei2022finetuned} which refines the model's ability to follow user instructions and engage in a conversational manner \citep{weber-etal-2024-investigating, zhang2025instructiontuninglargelanguage}. This process involves fine-tuning an LLM on an extensive dataset, which typically consists of instruction-answer pairs. By training on these pairs, the model learns to generate appropriate responses to user instructions, resulting in more anticipated and manageable outputs \citep{zhang2025instructiontuninglargelanguage}. A notable benefit of this approach is that it can enhance the model's ability to generalize and respond to previously unseen instructions \citep{nayak-etal-2024-learning, NEURIPS2023_ac662d74, sanh2022multitask}.

A significant issue at present is the scarcity of open-source, instruction tuning datasets for underrepresented languages. Existing datasets are predominantly in English, marginalizing other languages and leading to inferior model performance and higher deployment costs in these settings \citep{weber-etal-2024-investigating}. In the context of this work, "low-resource" refers to a scarcity of NLP tooling, open-source datasets, and specialized instruction-tuning resources. Translating an existing English instruction set, or pairing target-language content with a higher-resource language both depend on resources that are themselves scarce in these settings \citep{koksal-etal-2025-muri}. Luxembourgish, a West Germanic language spoken by about 600,000 people\footnote{\url{https://cursus.edu/en/23040/luxembourgish-at-its-best}, as of 11.08.2026} primarily in Luxembourg, exemplifies this challenge as adequate training data is scarce. 

In this paper, we introduce LuxIT, a monolingual instruction tuning dataset in Luxembourgish, synthesized from articles of the news website RTL\footnote{\url{https://www.rtl.lu/}} and Wikipedia entries. Following the work of \citet{lothritz-etal-2026-testing}, we select DeepSeek-R1-0528, as the top-performing model for Luxembourgish generation and comprehension to synthetically produce a final dataset of 227,507 high-quality instruction-answer pairs. To validate the quality of our dataset, we assess the Luxembourgish capabilities of several LLMs before and after they have been fine-tuned on LuxIT. 

To our knowledge, LuxIT is the first Luxembourgish instruction tuning dataset generated entirely within the target language, requiring neither a machine translation system nor parallel data.

Our primary contributions are twofold: 
\begin{itemize}[itemsep=-2pt]
    \item We introduce LuxIT\footnote{We provide the  \href{https://huggingface.co/datasets/jvalline/LuxIT_wiki_subset}{Wikipedia subset}, \href{https://language-data-space.eu/catalogue/view-offer/94463ebd-95db-478e-a0cc-e2f585006529/}{RTL subset} and  \href{https://github.com/JulianValline/LuxIT/tree/main}{code}}, a synthetically generated instruction tuning dataset in Luxembourgish.
    \item We evaluate the utility of LuxIT by fine-tuning 14 widely-used small-scale LLMs ($\leq$ 15B parameters) on LuxIT and evaluating them on standardized Luxembourgish proficiency exams and five downstream NLP tasks, demonstrating improvements in language proficiency for 12 of 14 models and NLP task performance for 9 of 14 models.
\end{itemize}
To validate the pipeline's ability to produce high-quality data, we assess the generated outputs against four custom metrics (see Section \ref{Post-filtering}), quantifying the proportion of retained samples. We then address the following research questions: 
\begin{itemize}[leftmargin=0pt,itemsep=0pt]
\item[] \textbf{RQ1: How does fine-tuning on LuxIT affect performance on Luxembourgish language exams?}
\item[] \textbf{i) Does fine-tuning on LuxIT improve overall accuracy across CEFR proficiency levels?} We fine-tune 14 LLMs ($\leq$ 15B parameters) on LuxIT and benchmark their performance in Luxembourgish against their corresponding base instruct models, following the evaluation framework by \citet{lothritz-etal-2026-testing}. 
\item[] \textbf{ii) How does the impact vary across individual linguistic categories?} We examine the results from RQ1 further and analyze them over the individual categorical exam levels, akin to \citet{lothritz-etal-2026-testing}. 
\item[] \textbf{RQ2: Does fine-tuning on LuxIT improve performance on downstream Luxembourgish NLP tasks?} We benchmark our fine-tuned models against their respective base instruct models on 5 well-known NLP tasks in Luxembourgish.
\end{itemize}

\section{Related work}
\subsection{Multilingual instruction tuning datasets}
xP3 \citep{muennighoff-etal-2023-crosslingual} is a multilingual, human-crafted instruction tuning dataset, combining data from existing datasets, such as P3 \citep{sanh2022multitask} into a unified format across 46 languages. The same work additionally introduces xP3mt, by applying machine translation to the English prompt templates. Bactrian-X \citep{li2023bactrianxmultilingualreplicableinstructionfollowing} machine-translates English instructions from Alpaca \citep{taori2023stanford} and Dolly \citep{conover2023dolly} into 51 other languages. Responses are then generated by GPT-3.5-Turbo. LIMA \citep{NEURIPS2023_ac662d74} is a fine-tuned version of Llama \citep{touvron2023llamaopenefficientfoundation}, trained on 1000 attentively selected instruction-answer pairs. \citet{weber-etal-2024-investigating} extend this resource to Lima-X by translating the LIMA instructions into 4 languages. Comparative overviews of these and related resources are provided by \citet{zhang2025instructiontuninglargelanguage} and \citet{weber-etal-2024-investigating}. However, none of these resources covers Luxembourgish.

\subsection{Luxembourgish language resources}
Recently, there have been various contributions to the creation of Luxembourgish language resources. \citet{lothritz-etal-2022-luxembert} developed a pre-training dataset in Luxembourgish through a data-augmentation technique where they partially translated text from German into Luxembourgish, resulting in LuxemBERT. \citet{plum-etal-2025-text} presented LuxGen, a benchmark for evaluating data generation in Luxembourgish and LuxT5, an mT5-based \citep{xue-etal-2021-mt5} text generation model in Luxembourgish, pre-trained on a German, French and Luxembourgish text corpus, where the latter is obtained through transfer learning from German and French. Most recently, \citet{philippy-etal-2025-luxembedder} introduced a hand-crafted cross-lingual dataset for training Luxembourgish sentence embedding models, which led to the LuxEmbedder model.

With the increasing ubiquity of LLMs, there is a need for a high-quality instruction tuning dataset in Luxembourgish to guide weaker models in improving their Luxembourgish capabilities. Two recent approaches generate Luxembourgish instruction data from native texts. \citet{koksal-etal-2025-muri} apply reverse instructions to entries from Wikipedia and CulturaX \citep{nguyen-etal-2024-culturax} across 200 languages, including Luxembourgish. Their pipeline however relies on machine translating the article into English and the generated instruction back into the target language. The cross-lingual dataset from \citet{philippy2025luxinstructcrosslingualinstructiontuning} demonstrates an effective strategy for generating instruction data in Luxembourgish without machine translation, pairing Luxembourgish with English, German, or French content. Yet, this approach is contingent on the availability of parallel seed data, which is often a bottleneck for low-resource languages.
Our work complements both approaches by performing generation directly in Luxembourgish, requiring neither machine translation nor parallel data.

\section{\textsc{LuxIT}}\label{Methodology}

\begin{figure}[ht]
  \centering
  \includegraphics[width=\columnwidth, keepaspectratio]{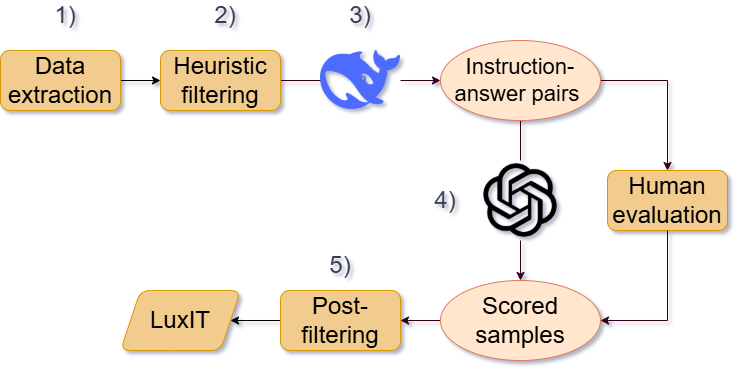}
  \caption{LuxIT Data Generation Pipeline.}
  \label{fig:data generation pipeline}
\end{figure}

This section details the methodology employed for the creation of our Luxembourgish instruction tuning dataset. The data generation process for LuxIT, illustrated in Figure \ref{fig:data generation pipeline} is executed as a five-step pipeline: The initial step (1) involves the extraction of raw data from two distinct sources (see Section \ref{Data}). In the second step (2), we subject the data to a series of heuristic filters (see Section~\ref{sec:filtering}). We subsequently provide the refined data to DeepSeek-R1-0528, which generates instruction-answer pairs in Luxembourgish (3). Following this, we implement an LLM-as-a-judge approach to evaluate the quality of our synthetic data (4). In parallel, a subset of these pairs undergoes manual human evaluation. In the final step, we apply a post-filtering process to remove all samples that received poor scores (5), resulting in the final LuxIT dataset.

\subsection{Data}\label{Data}
Our methodology utilizes two primary data sources: a complete dump of the Luxembourgish Wikipedia and a collection of all news articles and comments from RTL up to May 2024. We obtained the most recent dump of the Luxembourgish Wikipedia\footnote{\url{https://dumps.wikimedia.org/lbwiki/latest/}} on August 5th, 2025 and subsequently use Wikiextractor \citep{Wikiextractor2015} for extraction and cleaning. We convert the processed data from both sources into a JSON format for easier handling. Further details on the data structure are available in Appendix \ref{app: data_structure}. The initial collection comprises approximately 1.2 million RTL comments, 303,000 RTL news articles, and 80,000 Wikipedia articles, summing to approximately 1.6 million entries. We choose to exclude RTL user comments from our seed data due to their brevity and the potential for biased content, typographical mistakes, and grammatical inaccuracies. 

\subsection{Heuristic filtering}
\label{sec:filtering}
Before the generation phase, we implement a set of heuristic filters to remove low-quality samples. We retain only articles containing at least 750 characters. The threshold was set heuristically, to ensure each seed carries enough information. For Wikipedia articles, we exclude list articles, disambiguation pages, and pages marked as stubs. We filter out all non-Luxembourgish RTL news articles based on the language tags provided directly by RTL and apply additional filters, including boilerplate removal and semantic format filtering, which identifies keywords indicating a non-prose structure. This filtering process yields 16,558 Wikipedia articles and 100,340 RTL news articles.

\subsection{Instruction-Answer pair generation}

We employ DeepSeek-R1-0528\footnote{While \citet{lothritz-etal-2026-testing} used the original Deep-Seek-R1 in their experiments, we use the latest DeepSeek-R1-0528} to synthetically create instruction-answer pairs in Luxembourgish from the filtered RTL news and Wikipedia articles. We select all 16,558 Wikipedia articles and randomly sample 66,776 RTL news articles. We then prompt the model to generate three instruction-answer pairs for each seed, yielding a total of 245,624 synthetic pairs\footnote{There were some parsing errors for the generated json strings, hence we exclude those samples.}.

To ensure the generation of high-quality data, we engineer a comprehensive prompt designed to process information from both data sources effectively. The prompt instructs the model to embed all necessary context within the instruction to facilitate a self-contained answer. For news articles, we direct the model to handle temporal context with care. We also guide the model to produce a diverse set of instruction types, including summarization, question answering, information extraction, and explanation, where the latter task consists of providing a justification or reasoning alongside the response. For summarization tasks specifically, the model is required to include the original source text within the instruction itself. The complete data generation prompt is available in Appendix \ref{app: data_gen_prompt}.

The final LuxIT dataset is structured with four columns: \texttt{instruction}, \texttt{response}, \texttt{messages} and \texttt{source\_type} (RTL/Wiki). The \texttt{instruction} and \texttt{response} columns store the generated instructions and their corresponding answers. The \texttt{messages} column consolidates these two into ChatML format. We show the structure of our prompt template in Appendix \ref{app: prompt_template} and provide a representative sample of an instruction-answer pair from LuxIT in Table \ref{tab:LuxIT_sample} in Appendix \ref{app: LuxIT_sample}.

\subsection{LLM-as-a-judge scoring}\label{Post-filtering}

Following generation, we apply an LLM-as-a-judge approach for quality control using GPT-5-mini \citep{2025_OpenAI_GPT_5}. We select GPT-5-mini over GPT-4o-mini based on a comparison reported in Table \ref{tab:lux_benchmark} (Appendix \ref{app:llm_as_a_judge}) which shows that despite a slight underperformance at lower Common European Framework of Reference for Languages \citep{council2001common} (CEFR) levels, GPT-5-mini outperforms GPT-4o-mini by up to 24 points at higher levels (+9.7 pp macro average). A small-scale pilot comparing GPT-5 and GPT-5-mini judging found retention rates differing by only a few percent, justifying the cost-performance tradeoff. 

For the judging task, we prompt GPT-5-mini to assign scores to each pair based on four criteria: linguistic quality, factual accuracy, instruction adherence (or instruction following), and helpfulness \& relevance. Scoring is performed on the instruction–response pair in isolation. We use a three-point scoring system, with 1 representing poor quality and 3 indicating excellent quality. The linguistic quality score aims to check for the overall quality of the Luxembourgish language in the text, such as grammar, spelling and phrasing. Factual accuracy simply checks whether the answer factually answers the instruction without contradictions. Instruction adherence aims to verify whether the response correctly answers the instruction as intended and helpfulness relevance examines the general usefulness of the instruction. The full breakdown of our scoring system and the prompt used for this evaluation are provided in Appendices \ref{app: custom_scoring_metric} and \ref{app: post_filtering_prompt}. We retain only samples that achieve a score of at least 2 across all four metrics, discarding all others to form the final dataset.

\subsubsection{Human evaluation}

To validate our automated quality control, we conduct a manual human evaluation on a random subset of 100 generated samples. This sample size is selected to provide a meaningful preliminary audit of the dataset's quality while remaining feasible within the standard resource constraints of low-resource language research, where recruiting qualified native-speaker annotators is both challenging and cost-intensive. The evaluation was performed independently by three native Luxembourgish speakers (two male, one female; ages 26, 32, and 35) who hold academic backgrounds in computer science and history.

Each annotator evaluated the samples using the identical 3-point scoring rubric applied by the automated LLM judge. Annotators were instructed to assess each pair using only the information contained in the instruction and response, without consulting the seed article. All scoring disagreements are resolved through a strict majority vote to determine the final aggregated label for each sample.

The automated judge retains a higher proportion of samples (92.6\%) than the human evaluation (69.0\%), indicating a leniency bias in the LLM judge. Inter-annotator agreement (Fleiss' Kappa) is fair for linguistic quality ($\kappa=0.243$) but poor across the remaining metrics, including below-chance agreement for factual accuracy ($\kappa=-0.019$). This reflects the inherent difficulty of consistently evaluating synthetic Luxembourgish text and suggests that future annotation efforts would benefit from more granular rubrics and calibration sessions. Despite this low agreement, majority-vote aggregation over retained samples shows 94.2\% rated "Excellent" for factual accuracy, pointing to generally high quality while underscoring the subjectivity of the individual judgments. We report the full breakdown in Table \ref{tab:human_eval} in Appendix \ref{app: human_eval}.

\begin{table*}[ht]
\centering
\renewcommand{\arraystretch}{1.2}
\begin{tabular}{@{}lcccc@{}}
\hline
\textbf{Evaluation Metric} & \textbf{Exact Agreement} & \textbf{Cohen's $\kappa$} & \textbf{Human Mean (Std)} & \textbf{LLM Mean (Std)} \\
\hline
Linguistic Quality       & 57.0\% &  0.121 & 2.24 (0.57) & 2.19 (0.46) \\
Factual Accuracy         & 45.0\% & -0.058 & 2.84 (0.39) & 2.40 (0.65) \\
Instruction Adherence    & 83.0\% &  0.224 & 2.93 (0.26) & 2.80 (0.45) \\
Helpfulness Relevance & 51.0\% &  0.076 & 2.29 (0.88) & 2.74 (0.44) \\
\hline
\end{tabular}
\caption{Comparison of human and LLM evaluator (GPT-5-mini) agreement across four quality metrics on a random subset of 100 samples. Exact Agreement reports the percentage of matching scores, Cohen's $\kappa$ measures
inter-rater reliability, and Mean (Std) values are on a 1--3 scale.}
\label{tab:evaluation_metrics_human_judge}
\end{table*}

To directly evaluate the alignment between our automated judge and human insight, Table \ref{tab:evaluation_metrics_human_judge} presents a per-item agreement analysis across the 100-sample subset. Exact agreement is low and Cohen’s Kappa reveals poor reliability overall, highlighted by a negative coefficient for Factual Accuracy ($\kappa = -0.058$). This below-chance agreement is consistent with a mismatch visible in the metric means. For Factual Accuracy, human reviewers heavily favored perfect marks, resulting in a highly skewed mean of 2.84. Conversely, the LLM judge was significantly more conservative, frequently downgrading those same pairs to acceptable, leading to a lower mean of 2.40. By contrast, a reverse trend is visible in Helpfulness Relevance, where the LLM displays a leniency bias, overestimating value compared to humans (2.74 vs. 2.29).

\subsection{Post-filtering}\label{sec: post_filtering}

After filtering out all samples with at least one score of 1, the final version of our dataset contains 227,507 instruction-answer pairs in Luxembourgish. We provide a detailed overview in Table \ref{tab:score_distribution} (Appendix \ref{app: dataset_quality}) and show the score distribution on LuxIT from our post-filtering step.
The original dataset consists of 245,624 instruction-answer pairs. We reject 18,117 samples with low scores, observing a retention rate of 92.62\%. Among all samples, 32,215 (13.12\%) have perfect scores (score 3) across all score types. We observe that \texttt{instruction adherence} scores best among all metrics, with 223,065 ( 90.8\%) obtaining a score of 3.
Furthermore, among the rejected samples, \texttt{factual accuracy} has the highest rejection rate, with 11,745 entries having a score of 1. For the rejection rate, we observe 6,924, 11,745, 1,039 and 246 entries of score 1 (lowest) for linguistic quality, factual accuracy, instruction adherence and helpfulness relevance respectively.

\subsection{Dataset composition}

To characterize the task-type distribution of LuxIT and evaluate whether our open-ended generation prompt produces balanced coverage, we conduct a classification audit. While categorizing the entire 227,507-pair dataset is computationally prohibitive, we sample 100 random pairs and classify their instruction types using GPT-5.5 \cite{2026gpt55SystemCard}. The distribution of the sampled dataset is as follows: Summarization (32\%), Information Extraction (30\%), Question Answering (23\%), and Explanation (15\%). These results indicate that the generation prompt yields a relatively balanced mix of task types, though it exhibits a slight skew toward summarization and extraction tasks.
\section{Experimental setup}\label{sec:experimental_setup}

In this section, we outline and break down our experiments in detail.

\subsection{Fine-tuning on LuxIT}

We fine-tune a variety of small models ($\leq$15B parameters) on LuxIT. We split the dataset into train, evaluation and test split with ratios 0.92, 0.05 and 0.03 respectively, training on 209,307 and evaluating on 11,375 samples. We select the following models: Llama-3.1-8B-Instruct, Llama-3.2-1B-Instruct \citep{grattafiori2024llama3herdmodels}, Gemma-3-1B-IT, Gemma-3-12B-IT \citep{gemmateam2025gemma3technicalreport}, GLM-4-9B-0414 \citep{glm2024chatglm}, Qwen2.5-0.5B-Instruct, Qwen2.5-1.5B-Instruct, Qwen2.5-7B-Instruct \citep{qwen2025qwen25technicalreport}, 
Mistral-7B-Instruct-v0.3 \citep{jiang2023mistral7b}, Ministral-3-3B-Instruct-2512 \citep{liu2026ministral3}, Phi-4 \citep{abdin2024phi4technicalreport}, Apertus-8B-Instruct-2509 \citep{hernandez-cano-etal-2026-apertus}, EuroLLM-1.7B-Instruct \citep{MARTINS202553}, 
Olmo-3-7B-Instruct \citep{olmo2025olmo3}. The models are fine-tuned with LoRA \citep{hu2022lora} using the Unsloth\footnote{\url{https://docs.unsloth.ai/}} framework. Unless otherwise specified, we apply a standard configuration across models: a LoRA rank ($r$) of 16, a LoRA alpha ($\alpha$) of 32, a learning rate of $1 \times 10^{-4}$ with a cosine learning rate scheduler, and a warmup ratio of 0.03. Models are trained for 2 epochs using the AdamW 8-bit optimizer. Due to computational constraints,  hardware setups differ slightly between runs. A comprehensive table detailing the hyperparameters, exceptions, and GPU hardware used for each specific model can be found in Appendix \ref{app: fine_tuning_luxit}.

\subsection{Evaluation on language exams}\label{eval_lang_exams}
We investigate the impact of LuxIT on the Luxembourgish language proficiency of small LLMs, following the same approach proposed by \citet{lothritz-etal-2026-testing}.
After fine-tuning, we evaluate the LLMs on Luxembourgish language proficiency exams, instructing the models to solve exams consisting of multiple-choice questions, testing for vocabulary, grammar, reading comprehension, and conversational comprehension (done through transcripts). The language exams stem from the Luxembourgish language institute \textit{Institut National des Langues
Luxembourg (INLL)}\footnote{\url{https://www.inll.lu/en/}} and are divided into 6 skill levels defined in the CEFR, ranging from A1 (basic level) to C2 (native level), each exam level consisting of fill-in-the-blank and multiple choice questions  \citep{lothritz-etal-2026-testing}. 
The full dataset consists of 629 such questions, with slightly more than 100 questions per difficulty level and a nearly equal distribution of testing categories (vocabulary, grammar, reading comprehension, conversational comprehension).
We measure the models' performance using accuracy as a metric, akin to \citet{lothritz-etal-2026-testing}. We compare instruct models to our fine-tuned models.

The INLL exams are proprietary and not publicly accessible. This substantially reduces the risk of data contamination within our Wikipedia and RTL news seed corpora, ensuring that the models' performance reflects genuine language comprehension rather than test-set memorization.

\subsection{Evaluation on Luxembourgish downstream tasks}\label{sec: eval_downstream_tasks}

We evaluate our 14 fine-tuned models on 5 well-known NLP downstream tasks in Luxembourgish, namely Intent Classification (IC) \citep{10.1007/978-3-030-80599-9_32}, Winograd Natural Language Inference (WNLI) \cite{lothritz-etal-2022-luxembert}, Recognizing Textual Entailment (RTE), Sentiment Analysis and Stanford Sentiment Treebank (SST-2), taken from \citet{lothritz-etal-2023-comparing}. According to the authors, the data in these task-benchmarks was not taken from Wikipedia or RTL articles. The dataset consists of 3,823 samples. We report the class distribution in Table \ref{tab:task_distribution} in Appendix \ref{app: evaluation_downstream_tasks}. 
\section{Results}\label{sec:results}

\subsection{RQ1 i): Does fine-tuning on LuxIT improve overall accuracy across CEFR proficiency levels?}\label{rq1}

\begin{table*}[ht]
\centering
\resizebox{\textwidth}{!}{%
\begin{tabular}{lccccccc}
\toprule
\textbf{Model} & \textbf{A1} & \textbf{A2} & \textbf{B1} & \textbf{B2} & \textbf{C1} & \textbf{C2} & \textbf{Macro Avg} \\
& LuxIT ($\Delta$base) & LuxIT ($\Delta$base) & LuxIT ($\Delta$base) & LuxIT ($\Delta$base) & LuxIT ($\Delta$base) & LuxIT ($\Delta$base) & LuxIT ($\Delta$base) \\
\midrule
\multicolumn{8}{l}{\textit{$\sim$0.5--1.5B parameter models}} \\
qwen2.5-0.5b   & 20.2 (\textcolor{darkred}{$-$22.1})  & 21.2 (\textcolor{darkred}{$-$9.6})   & 12.6 (\textcolor{darkred}{$-$16.5})  & 23.7 (\textcolor{darkred}{$-$8.8})   & 15.4 (\textcolor{darkred}{$-$3.9})   & 6.9 (\textcolor{darkred}{$-$20.8})   & 16.7 (\textcolor{darkred}{$-$13.6})  \\
qwen2.5-1.5b   & 41.3 (\textcolor{darkgreen}{$+$1.0})  & 40.4 (\textcolor{darkgreen}{$+$1.0})  & 34.0 (\textcolor{darkgreen}{$+$10.7}) & 38.6 (\textcolor{darkgreen}{$+$15.8}) & 29.8 (\textcolor{darkgreen}{$+$1.0})  & 27.7 (\textcolor{darkgreen}{$+$6.9})  & 35.3 (\textcolor{darkgreen}{$+$6.0})  \\
llama-3.2-1b   & 36.5 (\textcolor{darkgreen}{$+$17.3}) & 32.7 (\textcolor{darkgreen}{$+$21.2}) & 24.3 (\textcolor{darkgreen}{$+$10.7}) & 26.3 (\textcolor{darkgreen}{$+$14.0}) & 21.2 (\textcolor{darkgreen}{$+$11.5}) & 22.8 (\textcolor{darkgreen}{$+$11.9}) & 27.3 (\textcolor{darkgreen}{$+$14.4}) \\
gemma-3-1b     & 23.1 (\textcolor{darkred}{$-$22.1})  & 21.2 (\textcolor{darkred}{$-$18.3})  & 13.6 (\textcolor{darkred}{$-$24.3})  & 10.5 (\textcolor{darkred}{$-$25.4})  & 7.7 (\textcolor{darkred}{$-$18.3})   & 5.9 (\textcolor{darkred}{$-$23.8})   & 13.7 (\textcolor{darkred}{$-$22.0})  \\
eurollm-1.7b   & 26.9 ($\pm$0.0)                       & 26.9 (\textcolor{darkgreen}{$+$3.8})  & 22.3 (\textcolor{darkgreen}{$+$2.9})  & 27.2 (\textcolor{darkgreen}{$+$7.9})  & 14.4 ($\pm$0.0)                       & 21.8 (\textcolor{darkgreen}{$+$8.9})  & 23.3 (\textcolor{darkgreen}{$+$3.9})  \\
ministral-3-3b & 61.5 (\textcolor{darkgreen}{$+$25.0}) & 48.1 (\textcolor{darkgreen}{$+$16.3}) & 45.6 (\textcolor{darkgreen}{$+$22.3}) & 48.2 (\textcolor{darkgreen}{$+$29.8}) & 38.5 (\textcolor{darkgreen}{$+$13.5}) & 38.6 (\textcolor{darkgreen}{$+$18.8}) & 46.8 (\textcolor{darkgreen}{$+$21.0}) \\
\midrule
\multicolumn{8}{l}{\textit{$\sim$7--9B parameter models}} \\
mistral-7b     & 51.9 (\textcolor{darkgreen}{$+$3.8})  & 45.2 (\textcolor{darkgreen}{$+$5.8})  & 41.7 (\textcolor{darkgreen}{$+$13.6}) & 44.7 (\textcolor{darkgreen}{$+$10.5}) & 29.8 (\textcolor{darkgreen}{$+$12.5}) & 26.7 (\textcolor{darkgreen}{$+$4.9})  & 40.0 (\textcolor{darkgreen}{$+$8.5})  \\
olmo-3-7b      & 46.2 (\textcolor{darkgreen}{$+$14.4}) & 48.1 (\textcolor{darkgreen}{$+$23.1}) & 38.8 (\textcolor{darkgreen}{$+$10.7}) & 33.3 (\textcolor{darkgreen}{$+$2.6})  & 35.6 (\textcolor{darkgreen}{$+$12.5}) & 22.8 (\textcolor{darkred}{$-$4.0})   & 37.5 (\textcolor{darkgreen}{$+$9.9})  \\
qwen2.5-7b     & 73.1 (\textcolor{darkgreen}{$+$21.2}) & 59.6 (\textcolor{darkgreen}{$+$14.4}) & 56.3 (\textcolor{darkgreen}{$+$16.5}) & 61.4 (\textcolor{darkgreen}{$+$15.8}) & 48.1 (\textcolor{darkgreen}{$+$14.4}) & 36.6 (\textcolor{darkred}{$-$1.0})   & 55.9 (\textcolor{darkgreen}{$+$13.6}) \\
llama-3.1-8b   & 49.0 ($\pm$0.0)                       & 46.2 (\textcolor{darkgreen}{$+$1.9})  & 41.7 (\textcolor{darkgreen}{$+$10.7}) & 40.4 (\textcolor{darkgreen}{$+$1.8})  & 39.4 (\textcolor{darkgreen}{$+$7.7})  & 25.7 (\textcolor{darkred}{$-$4.0})   & 40.4 (\textcolor{darkgreen}{$+$3.0})  \\
apertus-8b     & 56.7 (\textcolor{darkgreen}{$+$3.8})  & 51.0 (\textcolor{darkred}{$-$1.0})   & 48.5 (\textcolor{darkgreen}{$+$13.6}) & 46.5 (\textcolor{darkgreen}{$+$10.5}) & 36.5 (\textcolor{darkgreen}{$+$2.9})  & 32.7 (\textcolor{darkgreen}{$+$3.0})  & 45.3 (\textcolor{darkgreen}{$+$5.5})  \\
glm-4-9b  & 74.0 (\textcolor{darkgreen}{$+$11.5}) & 65.4 (\textcolor{darkgreen}{$+$2.9})  & 59.2 (\textcolor{darkgreen}{$+$11.6}) & 59.6 (\textcolor{darkgreen}{$+$7.0})  & 53.8 (\textcolor{darkgreen}{$+$7.7})  & 51.5 (\textcolor{darkgreen}{$+$8.9})  & 60.6 (\textcolor{darkgreen}{$+$8.3})  \\
\midrule
\multicolumn{8}{l}{\textit{$\sim$10--14B parameter models}} \\
gemma-3-12b    & 68.3 ($\pm$0.0)                       & 64.4 ($\pm$0.0)                       & 64.1 (\textcolor{darkgreen}{$+$8.7})  & 60.5 (\textcolor{darkgreen}{$+$3.5})  & 58.7 (\textcolor{darkgreen}{$+$5.8})  & 40.6 (\textcolor{darkred}{$-$3.0})   & 59.4 (\textcolor{darkgreen}{$+$2.5})  \\
phi-4          & 67.3 (\textcolor{darkgreen}{$+$7.7})  & 65.4 (\textcolor{darkgreen}{$+$16.3}) & 58.3 (\textcolor{darkgreen}{$+$20.4}) & 52.6 (\textcolor{darkgreen}{$+$7.0})  & 64.4 (\textcolor{darkgreen}{$+$24.0}) & 46.5 (\textcolor{darkgreen}{$+$9.9})  & 59.1 (\textcolor{darkgreen}{$+$14.2}) \\
\bottomrule
\end{tabular}}
\caption{Accuracy (\%) of LuxIT fine-tuned models on Luxembourgish language exams, with change relative to the base model (pre-fine-tuning instruct model) in brackets. \textcolor{darkgreen}{Green} indicates improvement over the base, \textcolor{darkred}{red} indicates degradation. $\Delta$base = macro-averaged accuracy of the LuxIT fine-tuned model minus that of the base model (higher is better).}
\label{tab:eval_results}
\end{table*}

Table \ref{tab:eval_results} shows the comparison between our fine-tuned models and their respective base models evaluated on Luxembourgish language exams, reporting total accuracy on all language exam levels. \\
Fine-tuning on LuxIT leads to a macro average accuracy increase in 12 out of 14 models, with macro average gains ranging from +2.5 percentage points (pp) (Gemma-3-12B-IT) to +21.0 pp (Ministral-3-3B-Instruct). On average, the models see a macro average accuracy increase of 5.37 pp. The effectiveness of instruction tuning on LuxIT is most evident for Ministral-3-3B-Instruct, with its macro average increasing from 25.8\% to 46.8\% (+21.0 pp). Similarly, for Llama-3.2-1B-Instruct, Phi-4 and Qwen2.5-7B-Instruct, the macro average increases by 14.4 pp, 14.2 pp and 13.6 pp respectively. For 2 of the 14 models tested, Gemma-3-1B and Qwen2.5-0.5B, we observe a degradation. We observe a decrease in macro average accuracy of -22.0 pp and -13.6 pp, respectively.
With regard to the individual language exam levels, we observe the most consistent increase for B1/B2 level across all models except our 2 overall decreasing models, Gemma-3-1B-IT and Qwen2.5-0.5B-Instruct. C2 level shows the most inconsistent gains, with 6 out of 14 models declining.

Out of all models, Gemma-3-12B-IT has the strongest baseline before fine-tuning, with a baseline macro average of 56.9\%. We identify GLM-4-9B-0414 as the top performing model in terms of absolute performance, with a final macro average of 60.6\%, followed by Gemma-3-12B-Instruct with an average of 59.4\%. Appendix \ref{app: rq1} contains a more detailed visual breakdown.

\subsection{RQ1 ii): How does the impact vary across individual linguistic categories?}

\begin{table}[t]
  \centering
  \setlength{\tabcolsep}{3pt}
  \small
  \resizebox{.48\textwidth}{!}{
  \begin{tabular}{@{}l
                  S[table-format=2.1]@{\;}l
                  S[table-format=2.1]@{\;}l
                  S[table-format=2.1]@{\;}l
                  S[table-format=2.1]@{\;}l@{}}
    \toprule
    \multirow{2}{*}{\textbf{Model}}
      & \multicolumn{2}{c}{\textbf{Vocabulary}}
      & \multicolumn{2}{c}{\textbf{Grammar}}
      & \multicolumn{2}{c}{\textbf{Reading}}
      & \multicolumn{2}{c}{\textbf{Conversation}} \\
    \cmidrule(lr){2-3}\cmidrule(lr){4-5}\cmidrule(lr){6-7}\cmidrule(lr){8-9}
      & {LuxIT} & {$\Delta$Base}
      & {LuxIT} & {$\Delta$Base}
      & {LuxIT} & {$\Delta$Base}
      & {LuxIT} & {$\Delta$Base} \\
    \midrule

    \multicolumn{9}{l}{\textit{$\sim$0.5--1.5B parameter models}} \\[1pt]

    qwen2.5-0.5b
      & 26.2 & \textcolor{darkred}{($-$4.1)}
      & 26.2 & \textcolor{darkred}{($-$5.1)}
      &  5.1 & \textcolor{darkred}{($-$25.6)}
      &  9.1 & \textcolor{darkred}{($-$19.9)} \\

    qwen2.5-1.5b
      & 36.8 & \textcolor{darkgreen}{($+$9.6)}
      & 33.3 & \textcolor{darkgreen}{($+$2.4)}
      & 33.3 & \textcolor{darkgreen}{($+$2.6)}
      & 38.0 & \textcolor{darkgreen}{($+$9.0)} \\

    llama-3.2-1b
      & 37.5 & \textcolor{darkgreen}{($+$24.4)}
      & 32.6 & \textcolor{darkgreen}{($+$8.8)}
      & 16.7 & \textcolor{darkgreen}{($+$7.7)}
      & 21.9 & \textcolor{darkgreen}{($+$16.8)} \\

    gemma-3-1b
      & 14.8 & \textcolor{darkred}{($-$24.6)}
      & 14.0 & \textcolor{darkred}{($-$24.1)}
      & 10.9 & \textcolor{darkred}{($-$21.8)}
      & 14.8 & \textcolor{darkred}{($-$17.4)} \\

    eurollm-1.7b
      & 29.0 & \textcolor{darkgreen}{($+$5.8)}
      & 29.9 & \textcolor{darkgreen}{($+$2.0)}
      & 14.1 & \textcolor{darkgreen}{($+$3.2)}
      & 19.3 & \textcolor{darkgreen}{($+$4.5)} \\

    ministral-3-3b
      & 54.7 & \textcolor{darkgreen}{($+$15.9)}
      & 44.9 & \textcolor{darkgreen}{($+$11.8)}
      & 43.6 & \textcolor{darkgreen}{($+$30.8)}
      & 44.5 & \textcolor{darkgreen}{($+$25.9)} \\

    \midrule
    \multicolumn{9}{l}{\textit{$\sim$7--9B parameter models}} \\[1pt]

    mistral-7b
      & 47.6 & \textcolor{darkgreen}{($+$20.1)}
      & 33.5 & \textcolor{darkgreen}{($+$1.9)}
      & 41.0 & \textcolor{darkgreen}{($+$8.3)}
      & 38.6 & \textcolor{darkgreen}{($+$4.6)} \\

    olmo-3-7b
      & 41.6 & \textcolor{darkgreen}{($+$14.0)}
      & 29.6 & \textcolor{darkgreen}{($+$1.0)}
      & 39.1 & \textcolor{darkgreen}{($+$13.5)}
      & 40.0 & \textcolor{darkgreen}{($+$10.9)} \\

    qwen2.5-7b
      & 65.8 & \textcolor{darkgreen}{($+$14.1)}
      & 40.0 & \textcolor{darkgreen}{($+$8.2)}
      & 59.0 & \textcolor{darkgreen}{($+$13.5)}
      & 59.3 & \textcolor{darkgreen}{($+$18.2)} \\

    llama-3.1-8b
      & 48.5 & \textcolor{darkgreen}{($+$7.9)}
      & 33.4 & \textcolor{darkgreen}{($+$1.8)}
      & 39.7 & \textcolor{darkgreen}{($+$4.5)}
      & 40.7 & \textcolor{darkred}{($-$1.8)} \\

    apertus-8b
      & 53.4 & \textcolor{darkred}{($-$2.2)}
      & 33.9 & \textcolor{darkgreen}{($+$8.5)}
      & 44.2 & \textcolor{darkgreen}{($+$4.5)}
      & 50.9 & \textcolor{darkgreen}{($+$11.0)} \\

    glm-4-9b
      & \bfseries 74.6 & \textcolor{darkgreen}{($+$3.1)}
      & 42.8 & \textcolor{darkgreen}{($+$7.1)}
      & 56.4 & \textcolor{darkgreen}{($+$7.7)}
      & 69.6 & \textcolor{darkgreen}{($+$15.5)} \\

    \midrule
    \multicolumn{9}{l}{\textit{$\sim$10--14B parameter models}} \\[1pt]

    gemma-3-12b
      & 70.5 & \textcolor{darkred}{($-$1.1)}
      & 41.1 & \textcolor{darkgreen}{($+$1.9)}
      & \bfseries 59.6 & \textcolor{darkgreen}{($+$4.5)}
      & 67.7 & \textcolor{darkgreen}{($+$5.1)} \\

    phi-4
      & 68.6 & \textcolor{darkgreen}{($+$18.3)}
      & \bfseries 43.4 & \textcolor{darkgreen}{($+$6.0)}
      & 54.5 & \textcolor{darkgreen}{($+$7.7)}
      & \bfseries 71.6 & \textcolor{darkgreen}{($+$25.8)} \\

    \bottomrule
  \end{tabular}}
  \caption{%
    Macro-averaged accuracy (\%) per skill category for each LuxIT fine-tuned model,
    together with the change over the respective base model.
    \textcolor{darkgreen}{Green} = improvement; \textcolor{darkred}{red} = degradation.
    $\Delta$Base = LuxIT accuracy minus base accuracy (higher is better).
  }
  \label{tab:categorical_results}
\end{table}

Table \ref{tab:categorical_results} shows macro-averaged accuracy changes across the four skill categories introduced in section \ref{eval_lang_exams}. The largest gains are observed in vocabulary and conversational comprehension, where multiple models improve substantially — notably Llama-3.2-1B-Instruct (+24.4 pp vocabulary) and Phi-4 (+25.8 pp conversational comprehension). Reading comprehension shows more variable results, with large gains for some models (Ministral-3-3B-Instruct: +30.8 pp) but modest improvements elsewhere. Grammar scores show the most modest changes, with gains rarely exceeding 10 pp, suggesting that the synthetic data is less effective at improving structural linguistic knowledge. The two degrading models (Qwen2.5-0.5B-Instruct and Gemma-3-1B-IT) show uniform drops across all four categories, consistent with their overall regression on language exams in Table \ref{tab:eval_results}.

\subsection{RQ2: Does fine-tuning on LuxIT improve performance on downstream Luxembourgish NLP tasks?}\label{sec: rq3}

\begin{figure}[ht]
    \centering
    \includegraphics[width=\columnwidth]{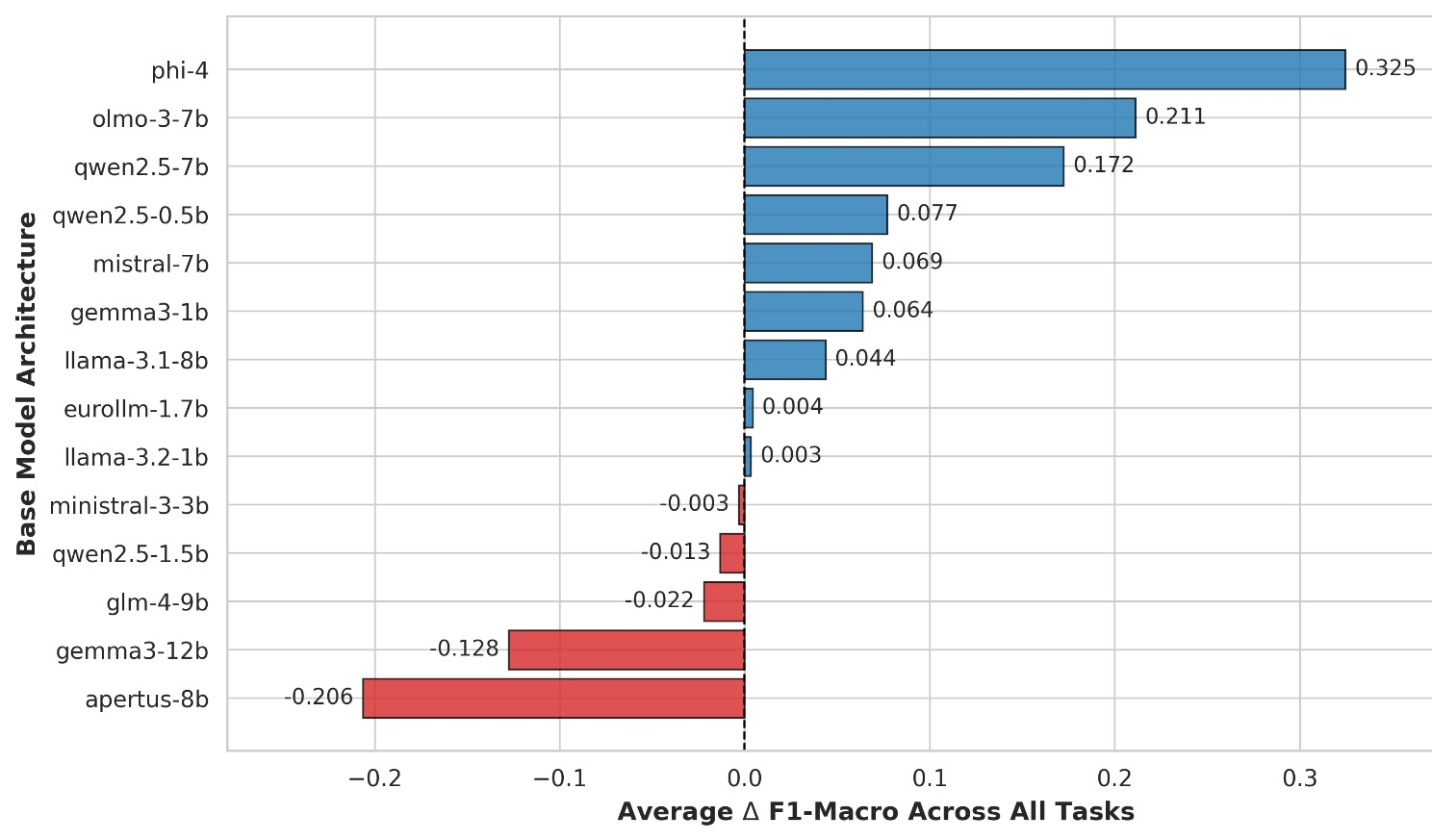}
    \caption{Average Impact of LuxIT Fine-Tuning per Model across all tasks.}
    \label{fig:barplot_downstream}
    \vspace{-18pt}
\end{figure}

Figure  \ref{fig:barplot_downstream} shows the average change in performance on the Luxembourgish NLP tasks (Section \ref{sec: eval_downstream_tasks}), comparing baseline and LuxIT fine-tuned models (see Appendix~\ref{app:rq3} for a breakdown of the results). Fine-tuning on LuxIT leads to improvements in Luxembourgish NLP task performance for 9 out of 14 models, as measured by macro-averaged F1 (Figure \ref{fig:barplot_downstream}) across intent classification, RTE, sentiment analysis, SST-2, and WNLI. The most substantial gains are observed for Phi-4 ($\Delta$F1-macro = +0.325), Olmo-3-7b (+0.211), and qwen2.5-7b (+0.172), while the largest regressions occur for Apertus-8b (-0.206) and Gemma-3-12b (-0.128). Models that improve on language exams (RQ1) do not systematically improve on NLP tasks, and vice versa with notable divergences for Ministral-3-3B, which shows the largest language exam gain (+21.0pp) alongside near-zero NLP task change (-0.003), and Gemma-3-1b, which shows a modest NLP task gain (+0.064) despite a substantial language exam regression (-22.0pp). 

It is to note that there are substantial outliers that influence the average performance change (see Appendix~\ref{app:rq3}, Figure~\ref{fig:heatmap_downstream_app}). Most notably, Apertus-8b performs poorly on the RTE task after fine-tuning on LuxIT, with a change in performance of -0.677 and Gemma3-12b's performance on IC changes by -0.522. On the other hand, we observe a considerable gain of +0.802  for Phi-4 in the SST task. While those outliers explain the extreme average changes we observe for Phi-4, Gemma3-12b, and Apertus-8b, it is unclear why those outliers occur.

\section{Discussion} 
\vspace{-5pt}
LuxIT represents an important step toward addressing the scarcity of instruction-tuning resources for Luxembourgish. Our pipeline effectively synthesizes a large dataset from monolingual seed data. Furthermore, our dataset composition audit demonstrates that the open-ended generation prompt yields a balanced coverage of diverse instruction types (summarization, extraction, Q\&A, and explanation). The human evaluation underscores the subjectivity of manually evaluating synthetic Luxembourgish text. We observe low human inter-annotator agreement (Table \ref{tab:human_eval}) alongside metric-specific biases between the automated judge and native speakers (Table \ref{tab:evaluation_metrics_human_judge}). While the LLM judge shows a leniency bias regarding Helpfulness Relevance, human evaluators are more forgiving regarding Factual Accuracy. This divergence highlights the need for more objective evaluation frameworks for future iterations.

Regarding language proficiency, fine-tuning generally enhanced formal Luxembourgish comprehension. However, the divergent outcomes among the smallest models suggest that fine-tuning receptivity depends on more than model size alone. Models with limited prior task alignment saw the most benefit, whereas regressions in other sub-1.5B models with stronger baselines likely point to catastrophic forgetting of pre-trained representations.

A deeper analysis of the language exam levels reveals that gains from LuxIT are not uniformly distributed across proficiency levels. Improvements were most consistent at intermediate proficiency levels (B1/B2) and in vocabulary and conversational comprehension. Contrarily, structural grammar and advanced native-like nuances (C2) proved difficult to improve. This suggests that the synthetic data, constrained by the generator model's capabilities and the formal nature of the seed corpora, effectively expands the models' lexicon but skews toward intermediate proficiency.

Furthermore, the limited correlation between language exam improvements and downstream NLP task performance highlights the specific nature of our dataset. The language exams feature standardized, well-formed Luxembourgish, while the NLP tasks contain noisy, non-standardized text. Because LuxIT is derived from formal Wikipedia and news articles, the fine-tuned models appear to align better with the language style of the formal exams. This divergence suggests that format alignment and linguistic improvement are separable effects that instruction tuning on LuxIT does not uniformly induce across all models.

Despite these nuances, in terms of absolute capability, fine-tuning medium-sized models on LuxIT successfully yielded highly proficient Luxembourgish models.

\section{Conclusion}

This work addresses the scarcity of instruction tuning resources for low-resource languages by introducing LuxIT, a high-quality dataset synthesized from monolingual Luxembourgish seed texts entirely within the target language. By fine-tuning fourteen small-scale LLMs, we demonstrate that LuxIT broadly improves formal language comprehension, particularly in expanding vocabulary and enhancing intermediate-level proficiency, even if performance on noisy, structured downstream NLP tasks remains less consistent. Ultimately, our findings validate that leveraging monolingual synthetic data is an effective and feasible strategy for enhancing LLM capabilities in low-resource linguistic settings. Future work will explore open-ended conversational evaluations and multi-turn data integration.
\section*{Limitations}

We acknowledge several limitations. First, LuxIT is derived from Wikipedia and RTL news articles and therefore skewed toward formal, standardized, and objective language registers. Consequently, LuxIT may lack the conversational flow, regional dialectal variations, or informal creative language styles of everyday Luxembourgish. Future data synthesis efforts should incorporate conversational transcripts or informal text. 

Second, our human evaluation, while crucial for validation, was conducted on a relatively small subset of 100 samples. A larger-scale human evaluation by multiple native speakers would lend greater statistical power to our quality claims and provide deeper insights into the dataset's strengths and weaknesses.

Third, we acknowledge the observed leniency bias of our single LLM judge (GPT-5-mini) and human evaluators. As shown by the low Exact Agreement and Cohen's Kappa scores (Table \ref{tab:evaluation_metrics_human_judge}), biases are metric-dependent. The LLM overestimates Helpfulness Relevance compared to humans, suggesting the final dataset may contain instructions that are marginally less useful than their automated scores imply. Furthermore, the human audit reveals a negative inter-annotator agreement among the native speakers themselves (Fleiss' $\kappa$ = -0.019, Table \ref{tab:human_eval}). Combined with their skewed leniency compared to the LLM (human mean of 2.84 vs. the LLM's 2.40), this indicates our evaluation guidelines allowed for highly subjective interpretations. This emphasizes the difficulty of manually evaluating synthetic text in Luxembourgish.  Future iterations should implement more granular rubrics and annotator calibration to improve human consistency, alongside stricter threshold prompting or multi-model LLM ensembles to enhance automated filtering. Similarly, our factual accuracy rubric should not be interpreted as a faithfulness metric. With the exception of the cases where the generation model reproduced the seed text within the instruction, neither the automated judge nor the human annotators were shown the source article. A response that is fluent and plausible but untrue with respect to the seed article would therefore not reliably be filtered out.

Fourth, while our approach is strictly monolingual in its seed data, the generation model, DeepSeek-R1-0528, is a multilingual model. Its internal knowledge is shaped by the vast multilingual corpus it was pre-trained on, which could subtly influence the style and structure of the generated Luxembourgish. More fundamentally, our pipeline requires a generator model already proficient in the target language.

Fifth, our downstream evaluation is restricted to multiple-choice language proficiency exams and classification tasks. While this provides a rigorous and objective measure of grammatical, lexical, and comprehension skills, it does not evaluate the models' open-ended text generation, multi-turn conversational abilities, or specific instruction-following nuances. Consequently, the true conversational proficiency of the fine-tuned models remains to be evaluated.

Finally, Apertus-8B-Instruct is constrained to a maximum sequence length of 512 tokens due to hardware limitations, versus 2048 tokens for all other models. As some LuxIT pairs may exceed this length, its results may underestimate the model's true capacity under standard training conditions.
\section*{Ethical considerations and licensing}

The Luxembourgish Wikipedia seed data is available under the CC BY-SA 4.0 license, and our use and distribution of RTL news articles is governed by a formal data-sharing agreement with RTL. As DeepSeek-R1-0528 places no restrictions on its outputs, we release the LuxIT Wikipedia subset under CC BY-SA 4.0 and the RTL subset under CC BY-NC 4.0. The INLL proficiency exams (Section \ref{eval_lang_exams}) are proprietary and not publicly distributed; we obtained explicit permission from INLL to use them solely for the automated evaluation of our models. The five downstream NLP benchmarks are used under their original academic terms.

Synthesizing data with LLMs risks propagating biases present in the seed texts or the generator model. We therefore excluded user-generated RTL comments, which are more prone to biased and toxic content, and applied heuristic filtering favoring formal, objective language.
\section*{Acknowledgments}

This research was supported by LLMs4EU, co- funded by the Digital Europe Programme under GA 101198470
, and by the BESSER project
through the Luxembourg National Research Fund
(FNR) PEARL programme under the grant agreement 16544475.
Our gratitude goes out to RTL for supplying the Luxembourgish news articles and to INLL for providing the Luxembourgish language exams. We thank Aline Fries and Aaron Conrardy for helping us with the human evaluation part and Bernhard Voggenberger for advising us on the fine-tuning experiments.

\bibliographystyle{acl_natbib}
\bibliography{anthology,latell}

@misc{zhang2025instructiontuninglargelanguage,
      title={Instruction Tuning for Large Language Models: A Survey}, 
      author={Shengyu Zhang and Linfeng Dong and Xiaoya Li and Sen Zhang and Xiaofei Sun and Shuhe Wang and others},
      year={2025},
      eprint={2308.10792},
      archivePrefix={arXiv},
      primaryClass={cs.CL},
      url={https://arxiv.org/abs/2308.10792}, 
      note={Preprint, arXiv:2308.10792}
}

@inproceedings{NEURIPS2020_1457c0d6,
 author = {Brown, Tom and Mann, Benjamin and Ryder, Nick and Subbiah, Melanie and Kaplan, Jared D and Dhariwal, Prafulla and others},
 booktitle = {Advances in Neural Information Processing Systems},
 editor = {H. Larochelle and M. Ranzato and R. Hadsell and M.F. Balcan and H. Lin},
 pages = {1877--1901},
 publisher = {Curran Associates, Inc.},
 title = {Language Models are Few-Shot Learners},
 url = {https://proceedings.neurips.cc/paper_files/paper/2020/file/1457c0d6bfcb4967418bfb8ac142f64a-Paper.pdf},
 volume = {33},
 year = {2020}
}

@misc{touvron2023llama2openfoundation,
      title={Llama 2: Open Foundation and Fine-Tuned Chat Models}, 
      author={Hugo Touvron and Louis Martin and Kevin Stone and Peter Albert and Amjad Almahairi and Yasmine Babaei and others},
      year={2023},
      eprint={2307.09288},
      archivePrefix={arXiv},
      primaryClass={cs.CL},
      url={https://arxiv.org/abs/2307.09288}, 
      note={Preprint, arXiv:2307.09288}
}

@inproceedings{NEURIPS2023_548a41b9,
 author = {Yin, Zhenfei and Wang, Jiong and Cao, Jianjian and Shi, Zhelun and Liu, Dingning and Li, Mukai and others},
 booktitle = {Advances in Neural Information Processing Systems},
 editor = {A. Oh and T. Naumann and A. Globerson and K. Saenko and M. Hardt and S. Levine},
 pages = {26650--26685},
 publisher = {Curran Associates, Inc.},
 title = {LAMM: Language-Assisted Multi-Modal Instruction-Tuning Dataset, Framework, and Benchmark},
 url = {https://proceedings.neurips.cc/paper_files/paper/2023/file/548a41b9cac6f50dccf7e63e9e1b1b9b-Paper-Datasets_and_Benchmarks.pdf},
 volume = {36},
 year = {2023}
}

@misc{peng2023instructiontuninggpt4,
      title={Instruction Tuning with GPT-4}, 
      author={Baolin Peng and Chunyuan Li and Pengcheng He and Michel Galley and Jianfeng Gao},
      year={2023},
      eprint={2304.03277},
      archivePrefix={arXiv},
      primaryClass={cs.CL},
      url={https://arxiv.org/abs/2304.03277}, 
      note={Preprint, arXiv:2304.03277}
}

@misc{li2023mimicitmultimodalincontextinstruction,
      title={MIMIC-IT: Multi-Modal In-Context Instruction Tuning}, 
      author={Bo Li and Yuanhan Zhang and Liangyu Chen and Jinghao Wang and Fanyi Pu and Jingkang Yang and Chunyuan Li and others},
      year={2023},
      eprint={2306.05425},
      archivePrefix={arXiv},
      primaryClass={cs.CV},
      url={https://arxiv.org/abs/2306.05425}, 
      note={Preprint, arXiv:2306.05425}
}

@inproceedings{NEURIPS2023_c2a8060f,
 author = {Shu, Manli and Wang, Jiongxiao and Zhu, Chen and Geiping, Jonas and Xiao, Chaowei and Goldstein, Tom},
 booktitle = {Advances in Neural Information Processing Systems},
 editor = {A. Oh and T. Naumann and A. Globerson and K. Saenko and M. Hardt and S. Levine},
 pages = {61836--61856},
 publisher = {Curran Associates, Inc.},
 title = {On the Exploitability of Instruction Tuning},
 url = {https://proceedings.neurips.cc/paper_files/paper/2023/file/c2a8060fd22744b38177d9e428a052e0-Paper-Conference.pdf},
 volume = {36},
 year = {2023}
}

@inproceedings{weber-etal-2024-investigating,
    title = "Investigating Multilingual Instruction-Tuning: Do Polyglot Models Demand for Multilingual Instructions?",
    author = "Weber, Alexander Arno  and
      Thellmann, Klaudia  and
      Ebert, Jan  and
      Flores-Herr, Nicolas  and
      Lehmann, Jens  and
      Fromm, Michael  and
      others",
    editor = "Al-Onaizan, Yaser  and
      Bansal, Mohit  and
      Chen, Yun-Nung",
    booktitle = "Proceedings of the 2024 Conference on Empirical Methods in Natural Language Processing",
    month = nov,
    year = "2024",
    address = "Miami, Florida, USA",
    publisher = "Association for Computational Linguistics",
    url = "https://aclanthology.org/2024.emnlp-main.1159/",
    doi = "10.18653/v1/2024.emnlp-main.1159",
    pages = "20829--20855"
}

@inproceedings{NEURIPS2023_ac662d74,
 author = {Zhou, Chunting and Liu, Pengfei and Xu, Puxin and Iyer, Srinivasan and Sun, Jiao and Mao, Yuning and others},
 booktitle = {Advances in Neural Information Processing Systems},
 doi = {10.52202/075280-2400},
 editor = {A. Oh and T. Naumann and A. Globerson and K. Saenko and M. Hardt and S. Levine},
 pages = {55006--55021},
 publisher = {Curran Associates, Inc.},
 title = {LIMA: Less Is More for Alignment},
 url = {https://proceedings.neurips.cc/paper_files/paper/2023/file/ac662d74829e4407ce1d126477f4a03a-Paper-Conference.pdf},
 volume = {36},
 year = {2023}
}

@inproceedings{nayak-etal-2024-learning,
    title = "Learning to Generate Instruction Tuning Datasets for Zero-Shot Task Adaptation",
    author = "Nayak, Nihal  and
      Nan, Yiyang  and
      Trost, Avi  and
      Bach, Stephen",
    editor = "Ku, Lun-Wei  and
      Martins, Andre  and
      Srikumar, Vivek",
    booktitle = "Findings of the Association for Computational Linguistics: ACL 2024",
    month = aug,
    year = "2024",
    address = "Bangkok, Thailand",
    publisher = "Association for Computational Linguistics",
    url = "https://aclanthology.org/2024.findings-acl.748/",
    doi = "10.18653/v1/2024.findings-acl.748",
    pages = "12585--12611"
}

@inproceedings{
sanh2022multitask,
title={Multitask Prompted Training Enables Zero-Shot Task Generalization},
author={Victor Sanh and Albert Webson and Colin Raffel and Stephen Bach and Lintang Sutawika and Zaid Alyafeai and others},
booktitle={International Conference on Learning Representations},
year={2022},
url={https://openreview.net/forum?id=9Vrb9D0WI4}
}

@inproceedings{
wei2022finetuned,
title={Finetuned Language Models are Zero-Shot Learners},
author={Jason Wei and Maarten Bosma and Vincent Zhao and Kelvin Guu and Adams Wei Yu and Brian Lester and others},
booktitle={International Conference on Learning Representations},
year={2022},
url={https://openreview.net/forum?id=gEZrGCozdqR}
}

@inproceedings{lothritz-etal-2026-testing,
    title = "Testing Low-Resource Language Support in {LLM}s Using Language Proficiency Exams: the Case of {L}uxembourgish",
    author = "Lothritz, Cedric  and
      Cabot, Jordi  and
      Bernardy, Laura",
    editor = "Demberg, Vera  and
      Inui, Kentaro  and
      Marquez, Llu{\'i}s",
    booktitle = "Findings of the {A}ssociation for {C}omputational {L}inguistics: {EACL} 2026",
    month = mar,
    year = "2026",
    address = "Rabat, Morocco",
    publisher = "Association for Computational Linguistics",
    url = "https://aclanthology.org/2026.findings-eacl.128/",
    doi = "10.18653/v1/2026.findings-eacl.128",
    pages = "2453--2476",
    ISBN = "979-8-89176-386-9"
}

@misc{2025_OpenAI_GPT_5,
  author = {OpenAI},
  title = {GPT-5 System Card},
  howpublished = {\url{https://openai.com/index/gpt-5-system-card/}},
  year = {2025},
  note = {Accessed on August 18th, 2025}
}

@inproceedings{muennighoff-etal-2023-crosslingual,
    title = "Crosslingual Generalization through Multitask Finetuning",
    author = "Muennighoff, Niklas  and
      Wang, Thomas  and
      Sutawika, Lintang  and
      Roberts, Adam  and
      Biderman, Stella  and
      Le Scao, Teven  and
      others",
    editor = "Rogers, Anna  and
      Boyd-Graber, Jordan  and
      Okazaki, Naoaki",
    booktitle = "Proceedings of the 61st Annual Meeting of the Association for Computational Linguistics (Volume 1: Long Papers)",
    month = jul,
    year = "2023",
    address = "Toronto, Canada",
    publisher = "Association for Computational Linguistics",
    url = "https://aclanthology.org/2023.acl-long.891/",
    doi = "10.18653/v1/2023.acl-long.891",
    pages = "15991--16111"
}

@misc{li2023bactrianxmultilingualreplicableinstructionfollowing,
      title={Bactrian-X: Multilingual Replicable Instruction-Following Models with Low-Rank Adaptation}, 
      author={Haonan Li and Fajri Koto and Minghao Wu and Alham Fikri Aji and Timothy Baldwin},
      year={2023},
      eprint={2305.15011},
      archivePrefix={arXiv},
      primaryClass={cs.CL},
      url={https://arxiv.org/abs/2305.15011}, 
      note={Preprint, arXiv:2305.15011}
}

@misc{touvron2023llamaopenefficientfoundation,
      title={LLaMA: Open and Efficient Foundation Language Models}, 
      author={Hugo Touvron and Thibaut Lavril and Gautier Izacard and Xavier Martinet and Marie-Anne Lachaux and Timothée Lacroix and others},
      year={2023},
      eprint={2302.13971},
      archivePrefix={arXiv},
      primaryClass={cs.CL},
      url={https://arxiv.org/abs/2302.13971}, 
      note={Preprint, arXiv:2302.13971}
}

@misc{taori2023stanford,
  title = {Alpaca: A Strong, Replicable Instruction-Following Model},
  author = {Rohan Taori and Ishaan Gulrajani and Tianyi Zhang and Yann Dubois and Xuechen Li and Carlos Guestrin and others},
  url = {https://crfm.stanford.edu/2023/03/13/alpaca.html},
  year = {2023},
  note = {Accessed on September 16, 2024},
}

@misc{grattafiori2024llama3herdmodels,
      title={The Llama 3 Herd of Models}, 
      author={Aaron Grattafiori and Abhimanyu Dubey and Abhinav Jauhri and Abhinav Pandey and Abhishek Kadian and Ahmad Al-Dahle and others},
      year={2024},
      eprint={2407.21783},
      archivePrefix={arXiv},
      primaryClass={cs.AI},
      url={https://arxiv.org/abs/2407.21783},
      note={Preprint, arXiv:2407.21783}
}

@misc{gemmateam2025gemma3technicalreport,
      title={Gemma 3 Technical Report}, 
      author={Gemma Team and Aishwarya Kamath and Johan Ferret and Shreya Pathak and Nino Vieillard and Ramona Merhej and others},
      year={2025},
      eprint={2503.19786},
      archivePrefix={arXiv},
      primaryClass={cs.CL},
      url={https://arxiv.org/abs/2503.19786},
      note={Preprint, arXiv:2503.19786}
}

@misc{glm2024chatglm,
      title={ChatGLM: A Family of Large Language Models from GLM-130B to GLM-4 All Tools}, 
      author={Team GLM and Aohan Zeng and Bin Xu and Bowen Wang and Chenhui Zhang and Da Yin and others},
      year={2024},
      eprint={2406.12793},
      archivePrefix={arXiv},
      primaryClass={id='cs.CL' full_name='Computation and Language' is_active=True alt_name='cmp-lg' in_archive='cs' is_general=False description='Covers natural language processing. Roughly includes material in ACM Subject Class I.2.7. Note that work on artificial languages (programming languages, logics, formal systems) that does not explicitly address natural-language issues broadly construed (natural-language processing, computational linguistics, speech, text retrieval, etc.) is not appropriate for this area.'},
      note = {Preprint, arXiv:2406.12793}
}

@inproceedings{
hu2022lora,
title={Lo{RA}: Low-Rank Adaptation of Large Language Models},
author={Edward J Hu and yelong shen and Phillip Wallis and Zeyuan Allen-Zhu and Yuanzhi Li and Shean Wang and others},
booktitle={International Conference on Learning Representations},
year={2022},
url={https://openreview.net/forum?id=nZeVKeeFYf9}
}

@inproceedings{lothritz-etal-2022-luxembert,
    title = "{L}uxem{BERT}: Simple and Practical Data Augmentation in Language Model Pre-Training for {L}uxembourgish",
    author = "Lothritz, Cedric  and
      Lebichot, Bertrand  and
      Allix, Kevin  and
      Veiber, Lisa  and
      Bissyande, Tegawende  and
      Klein, Jacques  and
      others",
    editor = "Calzolari, Nicoletta  and
      B{\'e}chet, Fr{\'e}d{\'e}ric  and
      Blache, Philippe  and
      Choukri, Khalid  and
      Cieri, Christopher  and
      Declerck, Thierry  and
      Goggi, Sara  and
      Isahara, Hitoshi  and
      Maegaard, Bente  and
      Mariani, Joseph  and
      Mazo, H{\'e}l{\`e}ne  and
      Odijk, Jan  and
      Piperidis, Stelios",
    booktitle = "Proceedings of the Thirteenth Language Resources and Evaluation Conference",
    month = jun,
    year = "2022",
    address = "Marseille, France",
    publisher = "European Language Resources Association",
    url = "https://aclanthology.org/2022.lrec-1.543/",
    pages = "5080--5089"
}

@inproceedings{plum-etal-2025-text,
    title = "Text Generation Models for {L}uxembourgish with Limited Data: A Balanced Multilingual Strategy",
    author = "Plum, Alistair  and
      Ranasinghe, Tharindu  and
      Purschke, Christoph",
    editor = "Scherrer, Yves  and
      Jauhiainen, Tommi  and
      Ljube{\v{s}}i{\'c}, Nikola  and
      Nakov, Preslav  and
      Tiedemann, Jorg  and
      Zampieri, Marcos",
    booktitle = "Proceedings of the 12th Workshop on NLP for Similar Languages, Varieties and Dialects",
    month = jan,
    year = "2025",
    address = "Abu Dhabi, UAE",
    publisher = "Association for Computational Linguistics",
    url = "https://aclanthology.org/2025.vardial-1.7/",
    pages = "93--104"
}

@inproceedings{philippy-etal-2025-luxembedder,
    title = "{L}ux{E}mbedder: A Cross-Lingual Approach to Enhanced {L}uxembourgish Sentence Embeddings",
    author = "Philippy, Fred  and
      Guo, Siwen  and
      Klein, Jacques  and
      Bissyande, Tegawende",
    editor = "Rambow, Owen  and
      Wanner, Leo  and
      Apidianaki, Marianna  and
      Al-Khalifa, Hend  and
      Eugenio, Barbara Di  and
      Schockaert, Steven",
    booktitle = "Proceedings of the 31st International Conference on Computational Linguistics",
    month = jan,
    year = "2025",
    address = "Abu Dhabi, UAE",
    publisher = "Association for Computational Linguistics",
    url = "https://aclanthology.org/2025.coling-main.753/",
    pages = "11369--11379"
}

@misc{philippy2025luxinstructcrosslingualinstructiontuning,
      title={LuxInstruct: A Cross-Lingual Instruction Tuning Dataset For Luxembourgish}, 
      author={Fred Philippy and Laura Bernardy and Siwen Guo and Jacques Klein and Tegawendé F. Bissyandé},
      year={2025},
      eprint={2510.07074},
      archivePrefix={arXiv},
      primaryClass={cs.CL},
      url={https://arxiv.org/abs/2510.07074},
      note={Preprint, arXiv:2510.07074}
}

@inproceedings{xue-etal-2021-mt5,
    title = "m{T}5: A Massively Multilingual Pre-trained Text-to-Text Transformer",
    author = "Xue, Linting  and
      Constant, Noah  and
      Roberts, Adam  and
      Kale, Mihir  and
      Al-Rfou, Rami  and
      Siddhant, Aditya  and
      others",
    editor = "Toutanova, Kristina  and
      Rumshisky, Anna  and
      Zettlemoyer, Luke  and
      Hakkani-Tur, Dilek  and
      Beltagy, Iz  and
      Bethard, Steven  and
      Cotterell, Ryan  and
      Chakraborty, Tanmoy  and
      Zhou, Yichao",
    booktitle = "Proceedings of the 2021 Conference of the North American Chapter of the Association for Computational Linguistics: Human Language Technologies",
    month = jun,
    year = "2021",
    address = "Online",
    publisher = "Association for Computational Linguistics",
    url = "https://aclanthology.org/2021.naacl-main.41/",
    doi = "10.18653/v1/2021.naacl-main.41",
    pages = "483--498"
}

@misc{qwen2025qwen25technicalreport,
      title={Qwen2.5 Technical Report}, 
      author={Qwen and : and An Yang and Baosong Yang and Beichen Zhang and Binyuan Hui and Bo Zheng and others},
      year={2025},
      eprint={2412.15115},
      archivePrefix={arXiv},
      primaryClass={cs.CL},
      url={https://arxiv.org/abs/2412.15115}, 
      note={Preprint, arXiv:2412.15115}
}

@misc{jiang2023mistral7b,
      title={Mistral 7B}, 
      author={Albert Q. Jiang and Alexandre Sablayrolles and Arthur Mensch and Chris Bamford and Devendra Singh Chaplot and Diego de las Casas and others},
      year={2023},
      eprint={2310.06825},
      archivePrefix={arXiv},
      primaryClass={cs.CL},
      url={https://arxiv.org/abs/2310.06825}, 
      note={Preprint, arXiv:2310.06825}
}

@misc{liu2026ministral3,
      title={Ministral 3}, 
      author={Alexander H. Liu and Kartik Khandelwal and Sandeep Subramanian and Victor Jouault and Abhinav Rastogi and Adrien Sadé and others},
      year={2026},
      eprint={2601.08584},
      archivePrefix={arXiv},
      primaryClass={cs.CL},
      url={https://arxiv.org/abs/2601.08584}, 
      note={Preprint, arXiv:2601.08584}
}

@inproceedings{hernandez-cano-etal-2026-apertus,
    title = "Apertus: Democratizing Open and Compliant {LLM}s for Global Language Environments",
    author = {Hern{\'a}ndez-Cano, Alejandro  and
      H{\"a}gele, Alexander  and
      Huang, Allen Hao  and
      Romanou, Angelika  and
      Solergibert, Antoni-Joan  and
      P{\'a}sztor, Barna  and
      others},
    editor = "Liakata, Maria  and
      Moreira, Viviane P.  and
      Zhang, Jiajun  and
      Jurgens, David",
    booktitle = "Proceedings of the 64th Annual Meeting of the {A}ssociation for {C}omputational {L}inguistics (Volume 1: Long Papers)",
    month = jul,
    year = "2026",
    address = "San Diego, California, United States",
    publisher = "Association for Computational Linguistics",
    url = "https://aclanthology.org/2026.acl-long.2172/",
    doi = "10.18653/v1/2026.acl-long.2172",
    pages = "46877--46955",
    ISBN = "979-8-89176-390-6"
}

@article{MARTINS202553,
title = {EuroLLM: Multilingual Language Models for Europe},
journal = {Procedia Computer Science},
volume = {255},
pages = {53-62},
year = {2025},
note = {Proceedings of the Second EuroHPC user day},
issn = {1877-0509},
doi = {https://doi.org/10.1016/j.procs.2025.02.260},
url = {https://www.sciencedirect.com/science/article/pii/S1877050925006210},
author = {Pedro Henrique Martins and Patrick Fernandes and João Alves and Nuno M. Guerreiro and Ricardo Rei and Duarte M. Alves and others}
}

@misc{olmo2025olmo3,
      title={Olmo 3}, 
      author={Team Olmo and : and Allyson Ettinger and Amanda Bertsch and Bailey Kuehl and David Graham and David Heineman and others},
      year={2025},
      eprint={2512.13961},
      archivePrefix={arXiv},
      primaryClass={cs.CL},
      url={https://arxiv.org/abs/2512.13961}, 
      note={Preprint, arXiv:2512.13961}
}

@misc{abdin2024phi4technicalreport,
      title={Phi-4 Technical Report}, 
      author={Marah Abdin and Jyoti Aneja and Harkirat Behl and Sébastien Bubeck and Ronen Eldan and Suriya Gunasekar and others},
      year={2024},
      eprint={2412.08905},
      archivePrefix={arXiv},
      primaryClass={cs.CL},
      url={https://arxiv.org/abs/2412.08905}, 
      note = {Preprint, arXiv:2412.08905}
}

@book{council2001common,
  title={{Common European framework of reference for languages: Learning, teaching, assessment}},
  author={{Council of Europe}},
  year={2001},
  publisher={Cambridge University Press},
  address   = {Cambridge, UK}
}

@inproceedings{lothritz-etal-2023-comparing,
    title = "Comparing Pre-Training Schemes for {L}uxembourgish {BERT} Models",
    author = "Lothritz, Cedric  and
      Ezzini, Saad  and
      Purschke, Christoph  and
      Bissyand{\'e}, Tegawend{\'e}  and
      Klein, Jacques  and
      Olariu, Isabella  and
      others",
    editor = "Georges, Munir  and
      Herygers, Aaricia  and
      Friedrich, Annemarie  and
      Roth, Benjamin",
    booktitle = "Proceedings of the 19th Conference on Natural Language Processing (KONVENS 2023)",
    month = sep,
    year = "2023",
    address = "Ingolstadt, Germany",
    publisher = "Association for Computational Lingustics",
    url = "https://aclanthology.org/2023.konvens-main.2/",
    pages = "17--27"
}

@misc{2026gpt55SystemCard,
  author = {OpenAI},
  title = {GPT-5.5 System Card},
  howpublished = {\url{https://deploymentsafety.openai.com/gpt-5-5/gpt-5-5.pdf}},
  year = {2026},
  note = {Accessed on March 6th, 2026}
}

@InProceedings{10.1007/978-3-030-80599-9_32,
author="Lothritz, Cedric
and Allix, Kevin
and Lebichot, Bertrand
and Veiber, Lisa
and Bissyand{\'e}, Tegawend{\'e} F.
and Klein, Jacques",
editor="M{\'e}tais, Elisabeth
and Meziane, Farid
and Horacek, Helmut
and Kapetanios, Epaminondas",
title="Comparing MultiLingual and Multiple MonoLingual Models for Intent Classification and Slot Filling",
booktitle="Natural Language Processing and Information Systems",
year="2021",
publisher="Springer International Publishing",
address="Cham",
pages="367--375",
isbn="978-3-030-80599-9"
}

@article{koksal-etal-2025-muri,
    title = "{MURI}: High-Quality Instruction Tuning Datasets for Low-Resource Languages via Reverse Instructions",
    author = {K{\"o}ksal, Abdullatif  and
      Thaler, Marion  and
      Imani, Ayyoob  and
      {\"U}st{\"u}n, Ahmet  and
      Korhonen, Anna  and
      Sch{\"u}tze, Hinrich},
    journal = "Transactions of the Association for Computational Linguistics",
    volume = "13",
    year = "2025",
    address = "Cambridge, MA",
    publisher = "MIT Press",
    url = "https://aclanthology.org/2025.tacl-1.48/",
    doi = "10.1162/tacl.a.18",
    pages = "1032--1055"
}

@inproceedings{nguyen-etal-2024-culturax,
    title = "{C}ultura{X}: A Cleaned, Enormous, and Multilingual Dataset for Large Language Models in 167 Languages",
    author = "Nguyen, Thuat  and
      Nguyen, Chien Van  and
      Lai, Viet Dac  and
      Man, Hieu  and
      Ngo, Nghia Trung  and
      Dernoncourt, Franck  and
      others",
    editor = "Calzolari, Nicoletta  and
      Kan, Min-Yen  and
      Hoste, Veronique  and
      Lenci, Alessandro  and
      Sakti, Sakriani  and
      Xue, Nianwen",
    booktitle = "Proceedings of the 2024 Joint International Conference on Computational Linguistics, Language Resources and Evaluation (LREC-COLING 2024)",
    month = may,
    year = "2024",
    address = "Torino, Italia",
    publisher = "ELRA and ICCL",
    url = "https://aclanthology.org/2024.lrec-main.377/",
    pages = "4226--4237"
}

@misc{conover2023dolly,
  title  = {Free {Dolly}: Introducing the World's First Truly Open
            Instruction-Tuned {LLM}},
  author = {Conover, Mike and Hayes, Matt and Mathur, Ankit and
            Xie, Jianwei and Wan, Jun and Shah, Sam and others},
  year   = {2023},
  url    = {https://www.databricks.com/blog/2023/04/12/dolly-first-open-commercially-viable-instruction-tuned-llm},
  note   = {Accessed on July 31, 2026},
}

@misc{Wikiextractor2015,
  author = {Giuseppe Attardi},
  title = {WikiExtractor},
  year = {2015},
  publisher = {GitHub},
  journal = {GitHub repository},
  howpublished = {\url{https://github.com/attardi/wikiextractor}}
}

\appendix

\appendix

\section{LuxIT}\label{app:LuxIT}

\subsection{Data structure}\label{app: data_structure}

Both data sources are contained in a JSON file. The following shows an example entry of an RTL news article\footnote{We do not show title, header and text for the RTL news article as we do not redistribute raw articles in this paper} and a Wikipedia article. We only extract \texttt{public\_date}, \texttt{title}, \texttt{header} and \texttt{text} from the RTL news article and extract \texttt{title} and \texttt{text} from the Wikipedia article.  

\begin{tcolorbox}[
    colback=blue!5,
    colframe=blue!40,
    title=RTL news article,
    fonttitle=\bfseries,
    width=\columnwidth,
    breakable=true,
    left=2pt, right=2pt, top=3pt, bottom=3pt,
]
\begin{lstlisting}[
    language=json,
    basicstyle=\footnotesize\ttfamily,
    breaklines=true,
    breakatwhitespace=true,
    keepspaces=true,
    columns=flexible,
    showstringspaces=false,
]
{
 'category_name': 'International',
 'category_id': 5,
 'article_id': 2190645,
 'type': 'news',
 'public_date': '2024-04-28 14:04:54',
 'title': '<article title>',
 'header': '<article header>',
 'text': '<article text>',
 'tags': ['international'],
 'text_id': 301779,
 'lang_id': 'lb'
}
\end{lstlisting}
\end{tcolorbox}

\begin{tcolorbox}[
    colback=blue!5,
    colframe=blue!40,
    title=Wikipedia article,
    fonttitle=\bfseries,
    width=\columnwidth,
    breakable=true,
    left=2pt, right=2pt, top=3pt, bottom=3pt,
]
\begin{lstlisting}[
    language=json,
    basicstyle=\footnotesize\ttfamily,
    breaklines=true,
    breakatwhitespace=true,
    keepspaces=true,
    columns=flexible,
    showstringspaces=false,
]
{
  "id": '25',
  "revid": '580', 
  "url": "https://lb.wikipedia.org/
    wiki?curid=25",
  "title": "Matthew Perry",
  "text": "De Matthew Langford Perry, gebuer den 19. August 1969 zu Williamstown am Massachusetts, a gestuerwen den 28. Oktober 2023 zu Los Angeles, war en US-amerikanesch-kanadesche Schauspiller, dee virun allem duerch seng Roll als Chandler Muriel Bing an der Televisiounsserie Friends bekannt ginn ass..."
}
\end{lstlisting}
\end{tcolorbox}

\subsection{Data generation model prompt}\label{app: data_gen_prompt}

The data generation is performed by DeepSeek-R1-0528. We instruct the model to return the data in JSON format, to ensure compatible formatting and to make it easily accessible. The prompt is formulated as follows:

\begin{tcblisting}{
    colback=yellow!10,
    colframe=orange!50,
    title=Data Generation Model Prompt,
    fonttitle=\bfseries,
    listing only,
    listing options={
        basicstyle=\footnotesize\ttfamily,  % Your preferred size
        breaklines=true,
        breakatwhitespace=true,
        keepspaces=true,
        columns=flexible,
        showstringspaces=false,
        xleftmargin=0pt,
        xrightmargin=0pt,
    },
    width=\columnwidth,
    breakable=true,                         % Allow breaking across columns
    left=2pt, right=2pt, top=3pt, bottom=3pt,
}
You are an expert in the Luxembourgish language tasked with creating high-quality synthetic training data for language models.

OBJECTIVE:
Generate 3 instruction-response pair(s) in authentic Luxembourgish based on the provided text. These pairs will be used for instruction fine-tuning of language models.

REQUIREMENTS:
1. LANGUAGE: All content MUST be in fluent, natural Luxembourgish
   - Use proper Luxembourgish grammar, spelling, and idioms
   - Avoid unnecessary German or French loan words
   - Ensure the language sounds natural to native speakers

2. QUALITY STANDARDS:
   - Instructions should be clear, specific, and answerable based on the provided text
   - Responses should be comprehensive, accurate, and well-structured
   - Include ALL necessary context in the instruction for a complete answer
   - If insufficient information exists, indicate that more details are needed

3. SUMMARIZATION INSTRUCTIONS:
   - When creating summarization tasks, ALWAYS include the original seed-text unchanged in the instruction for reference
   - The instruction should present the source text and ask for a summary
   - This ensures the training data contains both the source material and the summary

4. TEMPORAL CONTEXT:
   - When a date is provided, incorporate it appropriately
   - Add temporal context to maintain relevance when applicable
   - Consider whether dates belong in the instruction, response, or both

5. DIVERSITY:
   - Create varied types of instructions (e.g., summarization, Q&A, information extraction, explanation)
   - Vary complexity levels appropriately
   - Ensure each pair is unique and adds value

6. OUTPUT FORMAT:
   Return ONLY a valid JSON array with the following structure:
   [
     {{
       "instruction": "Clear instruction in Luxembourgish",
       "response": "Detailed response in Luxembourgish"
     }}
   ]

SOURCE TEXT:
{source_context}

Generate 3 high-quality instruction-response pair(s) based on the above text.
\end{tcblisting}

\subsection{LuxIT Sample}\label{app: LuxIT_sample}

Here we show a sample of LuxIT taken from the Wikipedia subset.

\begin{table}[ht]
\centering
\small
\begin{tabular}{p{0.44\columnwidth} p{0.44\columnwidth}}
\toprule
\textbf{Instruction} & \textbf{Response} \\
\midrule
Wien dierft bei de Gemengewalen wielen (aktiivt Walrecht), a wat sinn
d'Konditioune fir Net-L\"{e}tzebuerger? &
Fir ze wielen, muss een 18 Joer al sinn, biergerlech a politesch
Rechter hunn an zu L\"{e}tzebuerg wunnen. Net-L\"{e}tzebuerger mussen
hire Wunns\"{e}tz zu L\"{e}tzebuerg hunn, op der Gemeng an
d'Wielerl\"{e}scht ageschriwwen sinn, a bei Bierger vu L\"{a}nner
baussent EU/Schengen e g\"{u}ltegt Openthaltsdokument hunn. \\
\bottomrule
\end{tabular}
\caption{Example instruction-response pair from the LuxIT dataset, taken from the Wikipedia subset.}
\label{tab:LuxIT_sample}
\end{table}

\subsection{Prompt Template}\label{app: prompt_template}

Here we show the prompt template in more detail that unifies columns from LuxIT (Table \ref{tab:LuxIT_sample}). The \texttt{Instruction} and \texttt{Response} columns are brought together in the ChatML format:

\begin{tcolorbox}[
    colback=Violet!5,
    colframe=Violet!40,
    title=Prompt Template,
    fonttitle=\bfseries,
    width=\columnwidth,
    breakable=true,
    left=2pt, right=2pt, top=3pt, bottom=3pt,
]
\begin{lstlisting}[
    language=json,
    basicstyle=\footnotesize\ttfamily,
    breaklines=true,
    breakatwhitespace=true,
    keepspaces=true,
    columns=flexible,
    showstringspaces=false,
]
messages = 
[
    {
        "role": "user",
        "content": <instruction>
    },
    {
        "role": "assistant", 
        "content": <response>
    }
]
\end{lstlisting}
\end{tcolorbox}

\subsection{LLM-as-a-judge}\label{app:llm_as_a_judge}

Here we break down the reasoning for using GPT-5-mini as a judge (Section \ref{Post-filtering}). Table \ref{tab:lux_benchmark} shows the evaluation results from GPT-5 and GPT-5-mini compared to GPT-4o and GPT-4o-mini, where the accuracy scores for the latter 2 were taken from \citet{lothritz-etal-2026-testing} and we simply added the macro average. 

\begin{table}[h]
\centering
\renewcommand{\arraystretch}{1.15}
\resizebox{\columnwidth}{!}{%
\begin{tabular}{lcccccc|c}
\toprule
\textbf{Model} & \textbf{A1} & \textbf{A2} & \textbf{B1} & \textbf{B2}
               & \textbf{C1} & \textbf{C2} & \textbf{Avg} \\
\midrule
GPT-4o          & 92.3 & 85.6 & \textbf{78.6} & 84.2 & 74.0 & 60.4 & 79.2 \\
GPT-4o-mini     & 90.4 & 90.4 & 59.2          & 59.6 & 54.8 & 43.6 & 66.3 \\
\midrule
GPT-5           & \textbf{94.2} & \textbf{92.3} & \textbf{78.6} & \textbf{87.7}
                & \textbf{87.5} & \textbf{64.4} & \textbf{84.1} \\
\rowcolor{cyan!18}
GPT-5-mini      & 83.7 & 82.7 & 76.7 & 78.9 & 78.8 & 55.4 & 76.0 \\
\midrule
\textit{$\Delta$ (5-mini vs.\ 4o-mini)}
                & \textcolor{BrickRed}{$-$6.7} & \textcolor{BrickRed}{$-$7.7} &
\textcolor{OliveGreen}{$+$17.5} & \textcolor{OliveGreen}{$+$19.3} &
\textcolor{OliveGreen}{$+$24.0} & \textcolor{OliveGreen}{$+$11.8} &
\textcolor{OliveGreen}{$+$9.7} \\
\bottomrule
\end{tabular}%
}
\caption{Accuracy (\%) on Luxembourgish CEFR-level proficiency exams (A1--C2).
  \textbf{Bold} denotes the best score per column.
  \colorbox{cyan!20}{Shading} marks our chosen judge model.
  The $\Delta$ row shows the \textcolor{OliveGreen}{gain}/\textcolor{BrickRed}{degradation} of GPT-5-mini over GPT-4o-mini.}
\label{tab:lux_benchmark}
\end{table}

\subsection{Custom scoring metric}\label{app: custom_scoring_metric}

We fully break down our custom scoring metric introduced in section \ref{Post-filtering}: \\
\\
\textit{Linguistic Quality score}

\begin{itemize}[leftmargin=0pt]
    \item[] Score 1: Contains
significant grammatical errors,
spelling mistakes, or unnatural
phrasing in Luxembourgish. Text
that is actually German or French
instead of proper Luxembourgish
should receive this score.
    \item[] Score 2: Mostly
correct Luxembourgish, but has
minor errors or sounds slightly
robotic/unnatural. May mix in too
many loan words unnecessarily.
    \item[] Score 3: Fluent,
idiomatic, and grammatically
perfect Luxembourgish.
Natural-sounding text that a native speaker would produce.
\end{itemize}
\textit{Factual Accuracy score}

\begin{itemize}[leftmargin=0pt]
    \item[] Score 1: Contains
factual errors that contradict
the source text or general
knowledge.
    \item[] Score 2: Mostly
accurate but might have minor
inaccuracies or omissions.
    \item[] Score 3: Completely
accurate according to the source
text and factual knowledge.
\end{itemize}
\textit{Instruction Adherence score}

\begin{itemize}[leftmargin=0pt]
    \item[] Score 1: Fails to
follow the core instruction
(e.g., provides a summary when
asked for a list).
    \item[] Score 2: Follows the main instruction but
misses a constraint (e.g., writes
4 bullet points when asked for 3,
wrong format, or incorrect tone).
    \item[] Score 3: Perfectly follows all parts of
the instruction, including
constraints like length, format,
and tone.
\end{itemize}
\textit{Helpfulness Relevance score}

\begin{itemize}[leftmargin=0pt]
    \item[] Score 1: The
instruction is nonsensical,
irrelevant to any reasonable
context, or the response is
unhelpful/off-topic.
    \item[] Score 2: The
instruction is plausible but not
very insightful or creative. The
response addresses the
instruction but in a basic way.
    \item[] Score 3: A
genuinely useful, interesting, or
creative instruction that elicits
a helpful, comprehensive response.
\end{itemize}

\subsection{Post-filtering model prompt}\label{app: post_filtering_prompt}

We post filter the synthetic data with GPT-5-mini. Again, we want the scores to be in JSON format. The prompt is as follows:

\begin{tcblisting}{
    colback=ForestGreen!10,
    colframe=ForestGreen!50,
    title=Post-Filtering Model Prompt,
    fonttitle=\bfseries,
    listing only,
    listing options={
        basicstyle=\footnotesize\ttfamily,  % Your preferred size
        breaklines=true,
        breakatwhitespace=true,
        keepspaces=true,
        columns=flexible,
        showstringspaces=false,
        xleftmargin=0pt,
        xrightmargin=0pt,
    },
    width=\columnwidth,
    breakable=true,                         % Allow breaking across columns
    left=2pt, right=2pt, top=3pt, bottom=3pt,
}
You are an expert evaluator of Luxembourgish text quality. Your task is to evaluate the following instruction-response pair written in Luxembourgish based on four specific criteria.

IMPORTANT: The texts below are in Luxembourgish. You must evaluate them as Luxembourgish texts, NOT as German, French, or any other language. Luxembourgish has its own distinct grammar, vocabulary, and spelling conventions.

INSTRUCTION (in Luxembourgish):
{instruction}

RESPONSE (in Luxembourgish):
{response}

EVALUATION CRITERIA:

1. linguistic_quality (Linguistic Quality):
   - Score 1 (Poor): Contains significant grammatical errors, spelling mistakes, or unnatural phrasing in Luxembourgish. Text that is actually German or French instead of proper Luxembourgish should receive this score.
   - Score 2 (Acceptable): Mostly correct Luxembourgish, but has minor errors or sounds slightly robotic/unnatural. May mix in too many loan words unnecessarily.
   - Score 3 (Excellent): Fluent, idiomatic, and grammatically perfect Luxembourgish. Natural-sounding text that a native speaker would produce.

2. factual_accuracy (Factual Accuracy):
   - Score 1 (Incorrect): Contains factual errors that contradict the source text or general knowledge.
   - Score 2 (Mostly Correct): Mostly accurate but might have minor inaccuracies or omissions.
   - Score 3 (Perfect): Completely accurate according to the source text and factual knowledge.

3. instruction_adherence (Instruction Following):
   - Score 1 (Not Followed): Fails to follow the core instruction (e.g., provides a summary when asked for a list).
   - Score 2 (Partially Followed): Follows the main instruction but misses a constraint (e.g., writes 4 bullet points when asked for 3, wrong format, or incorrect tone).
   - Score 3 (Fully Followed): Perfectly follows all parts of the instruction, including constraints like length, format, and tone.

4. helpfulness_relevance (Helpfulness and Relevance):
   - Score 1 (Not Helpful): The instruction is nonsensical, irrelevant to any reasonable context, or the response is unhelpful/off-topic.
   - Score 2 (Somewhat Helpful): The instruction is plausible but not very insightful or creative. The response addresses the instruction but in a basic way.
   - Score 3 (Very Helpful): A genuinely useful, interesting, or creative instruction that elicits a helpful, comprehensive response.

CRITICAL INSTRUCTIONS:
- Respond ONLY with a JSON object containing the four scores.
- Each score must be an integer: 1, 2, or 3.
- Do NOT include any explanations, comments, or additional text outside the JSON.
- Evaluate the text AS LUXEMBOURGISH, not as any other language.

JSON FORMAT:
{{
    "linguistic_quality": <score 1-3>,
    "factual_accuracy": <score 1-3>,
    "instruction_adherence": <score 1-3>,
    "helpfulness_relevance": <score 1-3>
}}
\end{tcblisting}

\subsection{Dataset Quality}\label{app: dataset_quality}

In Table \ref{tab:score_distribution} we provide detailed statistics on the LuxIT score distribution introduced in Section \ref{sec: post_filtering}.

\begin{table*}[h]
\centering
\setlength{\tabcolsep}{6pt}
\renewcommand{\arraystretch}{0.85}
\begin{tabular}{lrrrrr}
\toprule
\textbf{Score type} & \textbf{Score 1} & \textbf{Score 2} & \textbf{Score 3} & \textbf{Mean} & \textbf{Median} \\
\midrule
\multicolumn{6}{c}{Original Dataset (245,624 entries)} \\
\midrule
LQ & 6,924 (2.8\%)  & 189,858 (77.3\%) & 48,842 (19.9\%)  & 2.171 & 2.0 \\
FA & 11,745 (4.8\%) & 101,693 (41.4\%) & 132,186 (53.8\%) & 2.490 & 3.0 \\
IA & 1,039 (0.4\%)  & 21,520 (8.8\%)   & 223,065 (90.8\%) & 2.904 & 3.0 \\
HR & 246 (0.1\%)    & 44,755 (18.2\%)  & 200,623 (81.7\%) & 2.816 & 3.0 \\
\midrule
\multicolumn{6}{c}{Filtered Dataset (227,507 entries)} \\
\midrule
LQ & N/A & 180,075 (79.2\%) & 47,432 (20.8\%)  & 2.208 & 2.0 \\
FA & N/A & 97,068 (42.7\%)  & 130,439 (57.3\%) & 2.573 & 3.0 \\
IA & N/A & 16,343 (7.2\%)   & 211,164 (92.8\%) & 2.928 & 3.0 \\
HR & N/A & 29,305 (12.9\%)  & 198,202 (87.1\%) & 2.871 & 3.0 \\
\bottomrule
\end{tabular}
\caption{LuxIT Score Distribution. We report scores from post-filtering on LuxIT with scores 1 (low), 2 (acceptable) and 3 (excellent) for Linguistic Quality (LQ), Factual Accuracy (FA), Instruction Adherence (IA) and Helpfulness Relevance (HR). We show the score distribution on the original dataset compared with the filtered dataset after rejecting 18,117 samples with low scores.}
\label{tab:score_distribution}
\end{table*}

\subsubsection{Human Evaluation}\label{app: human_eval}

We provide details for our human evaluation results in Table \ref{tab:human_eval}.

\begin{table*}[h]
\centering
\resizebox{\textwidth}{!}{%
\begin{tabular}{lcccc}
\toprule
\textbf{Evaluation Metric} & \textbf{Fleiss' $\kappa$} & \textbf{Aggregated Mean} & \textbf{\% Acceptable (Score 2)} & \textbf{\% Excellent (Score 3)} \\
\midrule
Linguistic Quality       & 0.243  & 2.33 & 66.7\% & 33.3\% \\
Factual Accuracy         & -0.019 & 2.94 & 5.8\%  & 94.2\% \\
Instruction Adherence    & 0.102  & 2.93 & 7.2\%  & 92.8\% \\
Helpfulness Relevance & 0.090  & 2.80 & 20.3\% & 79.7\% \\
\bottomrule
\end{tabular}%
}
\caption{Human evaluation results on a random subset of 100 samples. Inter-annotator agreement is
reported using Fleiss' $\kappa$ across three independent evaluators. The final aggregated scores (Majority Vote)
are reported for the 69 retained samples (scores of 1 ``Low'' are discarded).}
\label{tab:human_eval}
\end{table*}
\section{Experimental Setup}\label{app:Experimental_setup}

Here we provide more details accompanying the Experimental Setup Section (Section \ref{sec:experimental_setup}).

\subsection{Fine-Tuning on LuxIT}\label{app: fine_tuning_luxit}

We provide more details for the hyperparameter settings in Table \ref{tab:hyperparameters}. (Introduced in Section \ref{sec:experimental_setup}). Several hyperparameters and configurations are held constant across all models unless otherwise specified.

All models are fine-tuned for 2 epochs using the 8-bit AdamW optimizer and a cosine learning rate scheduler. We consistently apply a warmup ratio of 0.03, a weight decay of 0.01, and keep the default value for max\_grad\_norm at 1.0. For the Low-Rank Adaptation (LoRA) setup, the rank ($r$) is set to 16, the scaling factor ($\alpha$) to 32, and the dropout rate to 0.05. Gradient checkpointing is enabled across all training runs to optimize memory consumption. During training, logging occurs every 200 steps, while evaluation and model checkpoint saving are performed every 500 steps. For the dataset, 5\% is strictly held out for validation and 3\% for testing across all runs.

Notably, while a maximum sequence length of 2048 tokens is utilized for the vast majority of the models, we constrain the Apertus-8B-Instruct model to a sequence length of 512 tokens due to hardware memory limitation requirements. Furthermore, all models trained via the Unsloth framework utilize a fixed random seed of 3407 to ensure reproducibility, whereas the Apertus-8B-Instruct model, which we train using standard Hugging Face pipelines, is initialized with a random seed of 42.

\begin{table*}[h]
\centering
\begin{tabular}{llccccl}
\toprule
\textbf{Model} & \textbf{GPU} & \textbf{4-bit Load} & \textbf{BS} & \textbf{GAS} & \textbf{LR} & \textbf{Target Modules} \\
\midrule
Gemma-3-12B-IT & RTX 4090 & Yes & 2 & 16 & $1 \times 10^{-4}$ & Gemma-specific\textsuperscript{1} \\
Olmo-3-7B-Instruct & RTX 4090 & No & 8 & 4 & $2 \times 10^{-4}$ & Standard\textsuperscript{2} \\
EuroLLM-1.7B-Instruct & RTX 4090 & No & 16 & 3 & $1 \times 10^{-4}$ & Standard\textsuperscript{2} \\
Gemma-3-1B-IT & RTX 4090 & No & 16 & 3 & $1 \times 10^{-4}$ & Gemma-specific\textsuperscript{1} \\
Apertus-8B-Instruct\textsuperscript{3} & RTX 4090 & No & 2 & 16 & $2 \times 10^{-4}$ & all-linear \\
Qwen2.5-7B-Instruct & V100 & No & 2 & 16 & $1 \times 10^{-4}$ & Standard\textsuperscript{2} \\
Qwen2.5-1.5B-Instruct & V100 & No & 2 & 16 & $2 \times 10^{-4}$ & Standard\textsuperscript{2} \\
Qwen2.5-0.5B-Instruct & V100 & No & 2 & 16 & $2 \times 10^{-4}$ & Standard\textsuperscript{2} \\
Phi-4 & V100 & Yes & 2 & 16 & $1 \times 10^{-4}$ & Standard\textsuperscript{2} \\
Mistral-7B-Instruct-v0.3 & V100 & No & 4 & 8 & $1 \times 10^{-4}$ & Standard\textsuperscript{2} \\
Ministral-3-3B & V100 & No & 4 & 10 & $2 \times 10^{-4}$ & Standard\textsuperscript{2} \\
Llama-3.2-1B-Instruct & V100 & No & 8 & 6 & $2 \times 10^{-4}$ & Standard\textsuperscript{2} \\
Llama-3.1-8B-Instruct & V100 & No & 2 & 16 & $1 \times 10^{-4}$ & Standard\textsuperscript{2} \\
GLM-4-9B-0414 & V100 & No & 2 & 16 & $1 \times 10^{-4}$ & Standard\textsuperscript{2} \\
\bottomrule
\end{tabular}
\caption{Hyperparameter Settings by Model. Constants across all models include 2 training epochs, AdamW 8-bit optimizer, cosine learning rate scheduler, a warmup ratio of 0.03, weight decay of 0.01, and LoRA parameters ($r=16,  \alpha=32$, dropout = $0.05$). BS = Train Batch Size per Device, GAS = Gradient Accumulation Steps, LR = Learning Rate. \textsuperscript{1}Gemma-specific: vision\_layers=False, language\_layers=True, attention\_modules=True, mlp\_modules=True. \textsuperscript{2}Standard: q\_proj, k\_proj, v\_proj, o\_proj, gate\_proj, up\_proj, down\_proj. \textsuperscript{3}Apertus-8B-Instruct was the only model fine-tuned utilizing standard Hugging Face pipelines rather than Unsloth.}
\label{tab:hyperparameters}
\end{table*}

\subsection{Evaluation on Luxembourgish downstream tasks}\label{app: evaluation_downstream_tasks}

We break down the dataset class distribution in Table \ref{tab:task_distribution} introduced in Section \ref{sec: eval_downstream_tasks}.

\begin{table}[H]
\centering
\small
\setlength{\tabcolsep}{4pt}
\begin{tabular}{@{}lr@{}}
\toprule
\textbf{Task} & \textbf{Samples} \\
\midrule
Intent Classification (IC) & 159 \\
Recognizing Textual Entailment (RTE) & 801 \\
Sentiment Analysis (SA) & 367 \\
Stanford Sentiment Treebank (SST-2) & 2,360 \\
Winograd NLI (WNLI) & 136 \\
\bottomrule
\end{tabular}
\caption{Sample distribution across Luxembourgish downstream tasks.}
\label{tab:task_distribution}
\end{table}
\section{Results}

Here we provide more details accompanying the Results section (Section \ref{sec:results}).

\subsection{RQ1 i)}\label{app: rq1}

We provide detailed visual comparison between our fine-tuned models and their respective baseline in Figure \ref{fig:all-model-comparisons} (Section \ref{rq1}).

\begin{figure*}[p]
    \centering
    \captionsetup[subfigure]{skip=2pt}
    \setlength{\abovecaptionskip}{6pt}
    \begin{subfigure}[t]{0.45\textwidth}
        \centering
        \includegraphics[width=\textwidth]{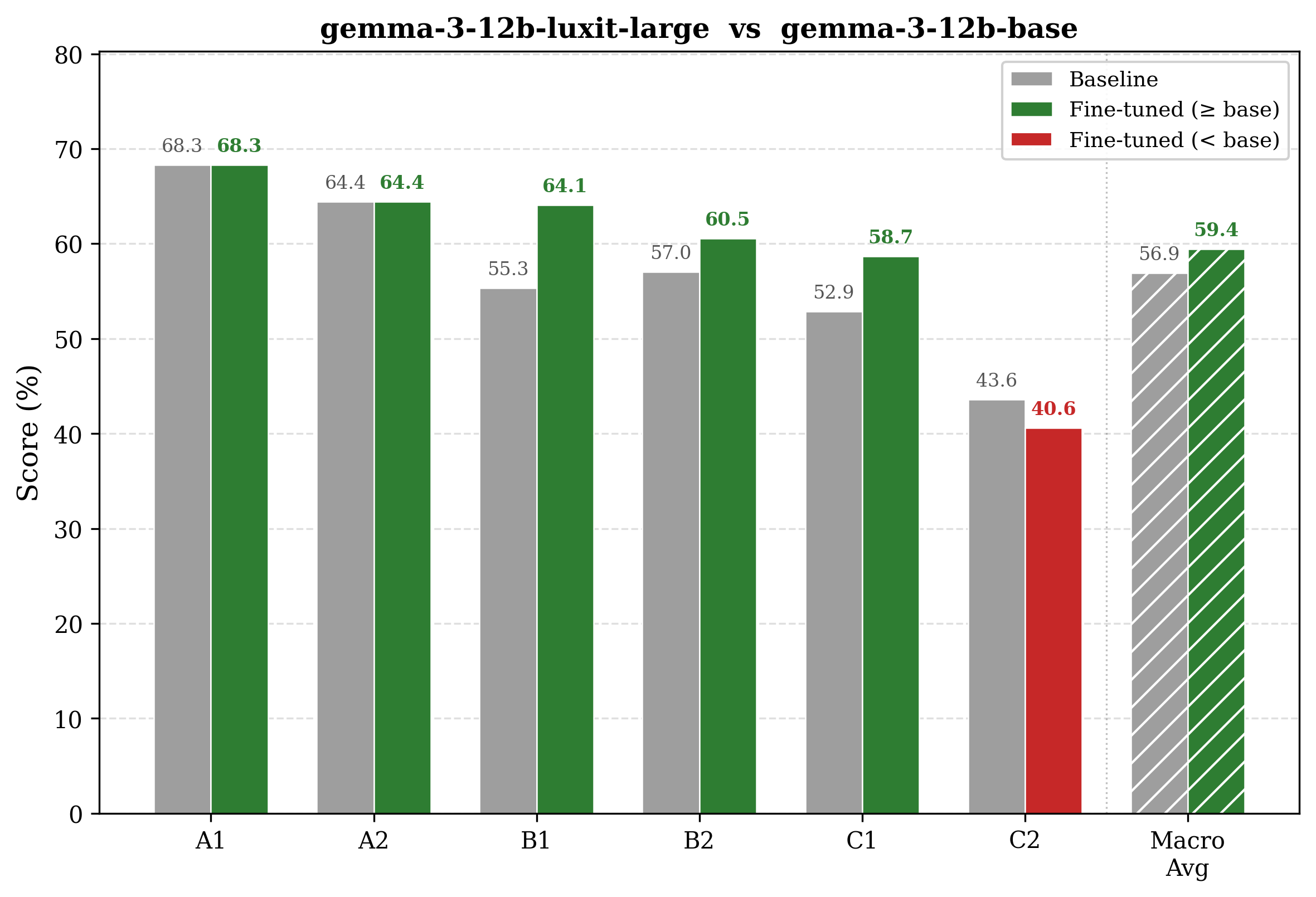}
        \caption{gemma-3-12b}
        \label{fig:gemma-3-12b}
    \end{subfigure}
    \hfill
    \begin{subfigure}[t]{0.45\textwidth}
        \centering
        \includegraphics[width=\textwidth]{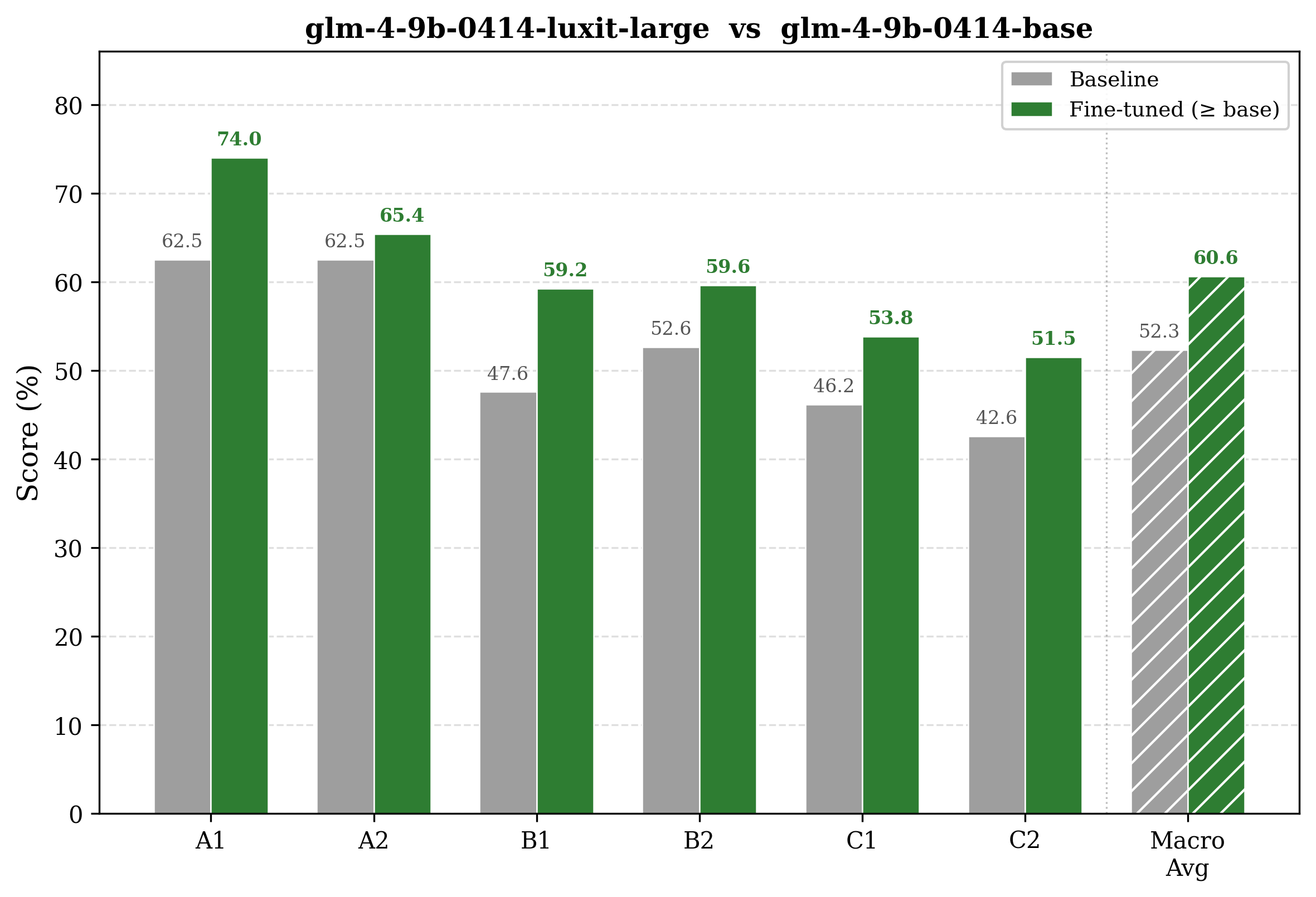}
        \caption{glm-4-9b-0414}
        \label{fig:glm-4-9b}
    \end{subfigure}
    \begin{subfigure}[t]{0.45\textwidth}
        \centering
        \includegraphics[width=\textwidth]{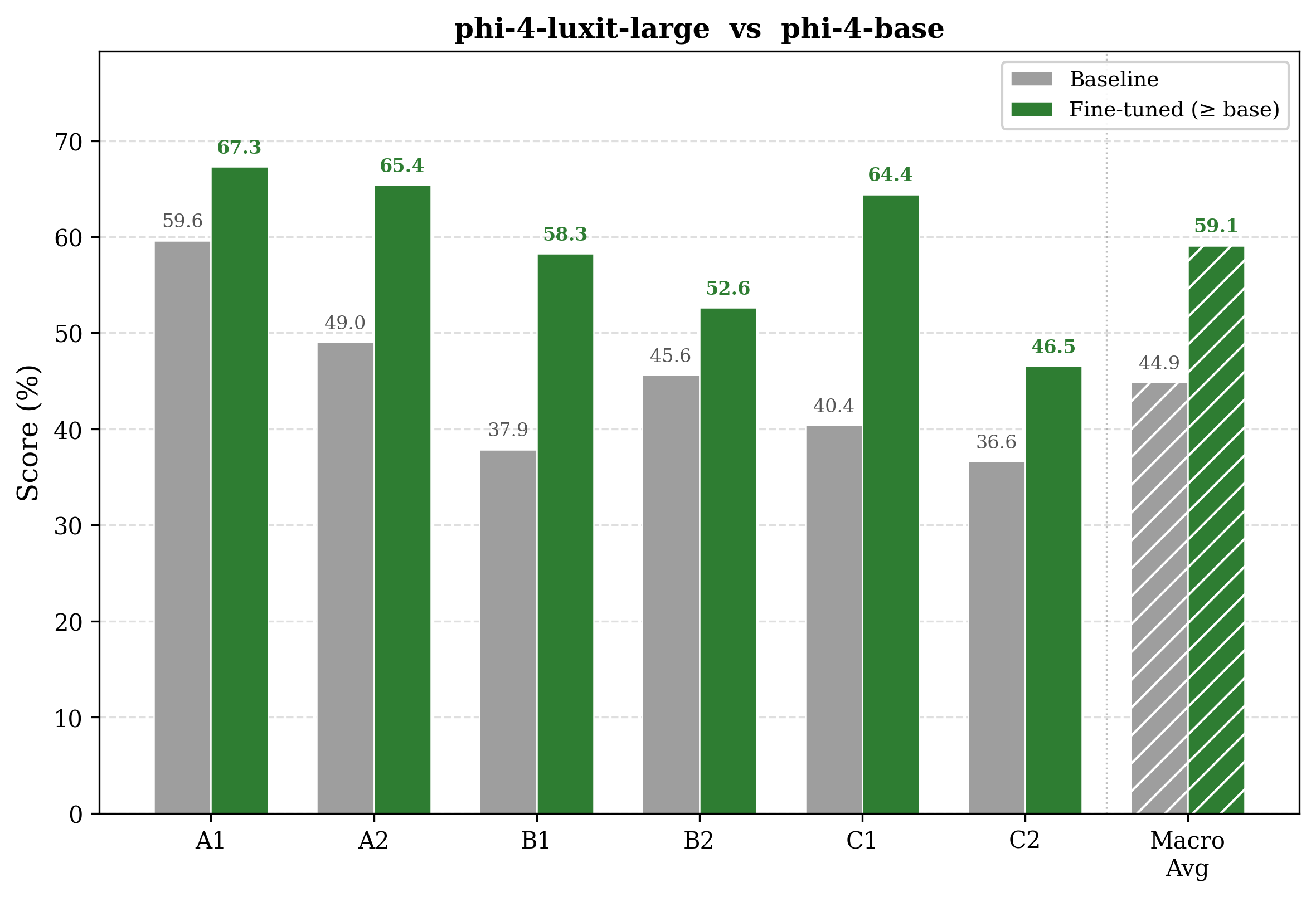}
        \caption{phi-4}
        \label{fig:phi-4}
    \end{subfigure}
    \hfill
    \begin{subfigure}[t]{0.45\textwidth}
        \centering
        \includegraphics[width=\textwidth]{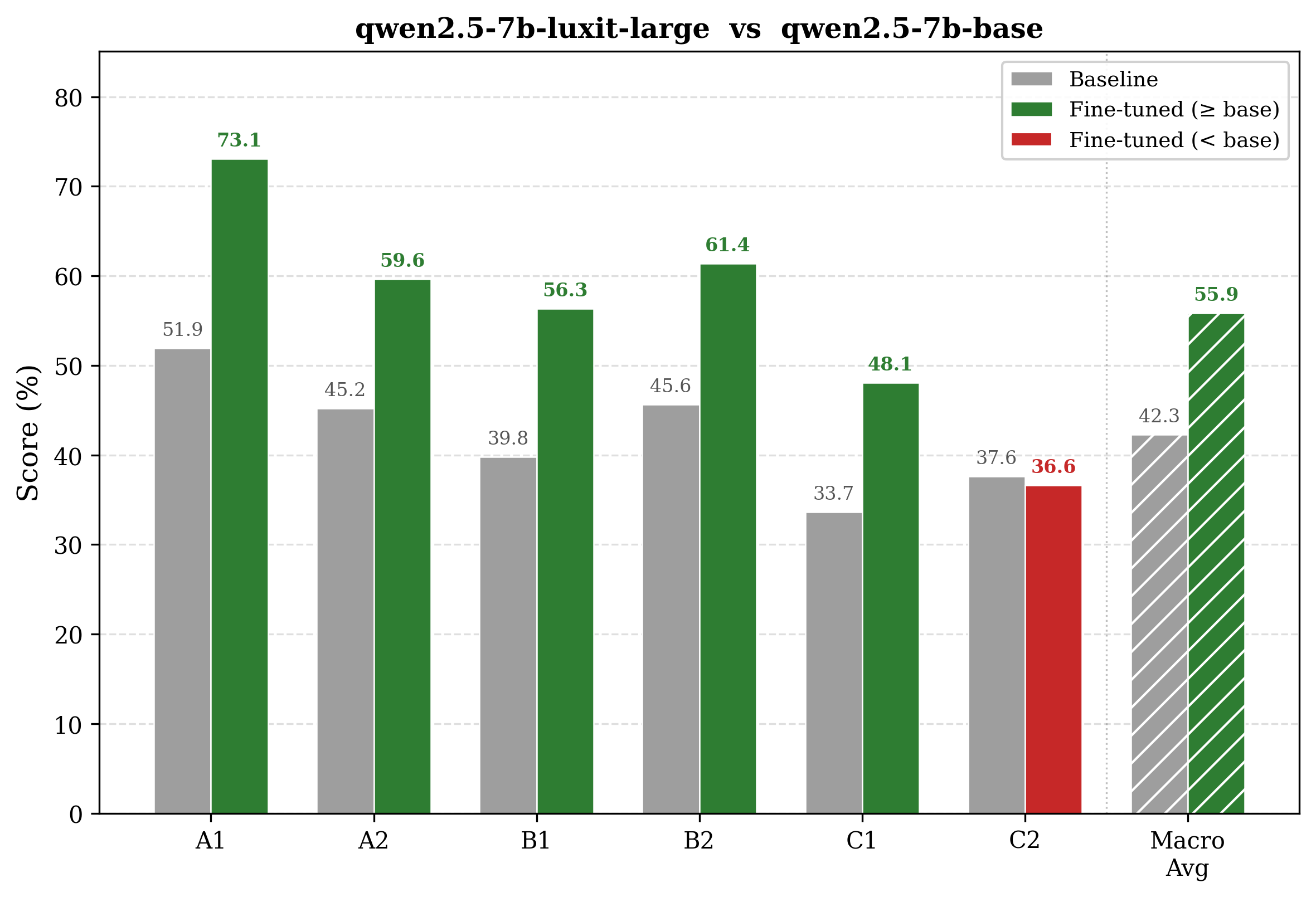}
        \caption{qwen2.5-7b}
        \label{fig:qwen2.5-7b}
    \end{subfigure}
    \begin{subfigure}[t]{0.45\textwidth}
        \centering
        \includegraphics[width=\textwidth]{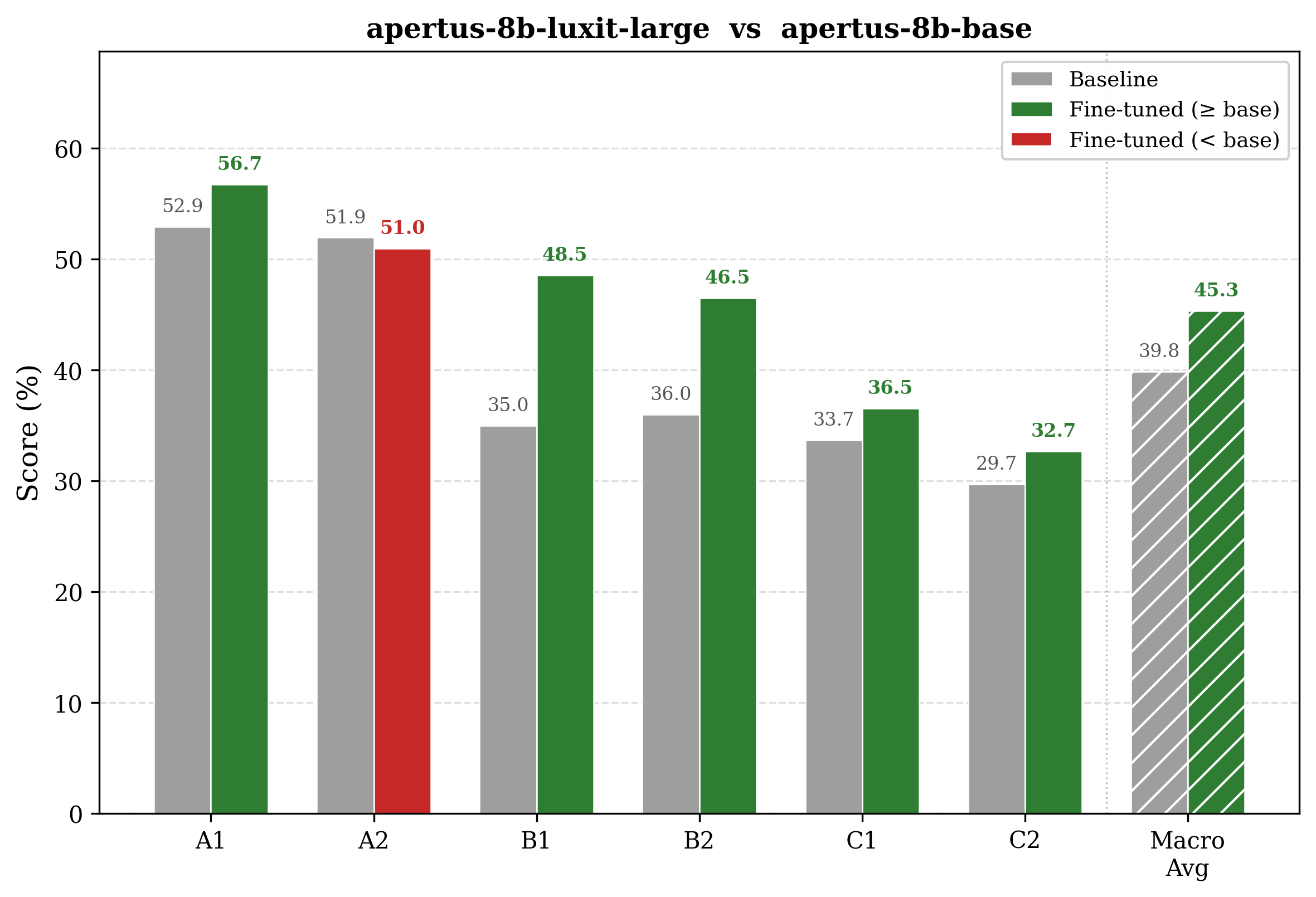}
        \caption{apertus-8b}
        \label{fig:apertus-8b}
    \end{subfigure}
    \hfill
    \begin{subfigure}[t]{0.45\textwidth}
        \centering
        \includegraphics[width=\textwidth]{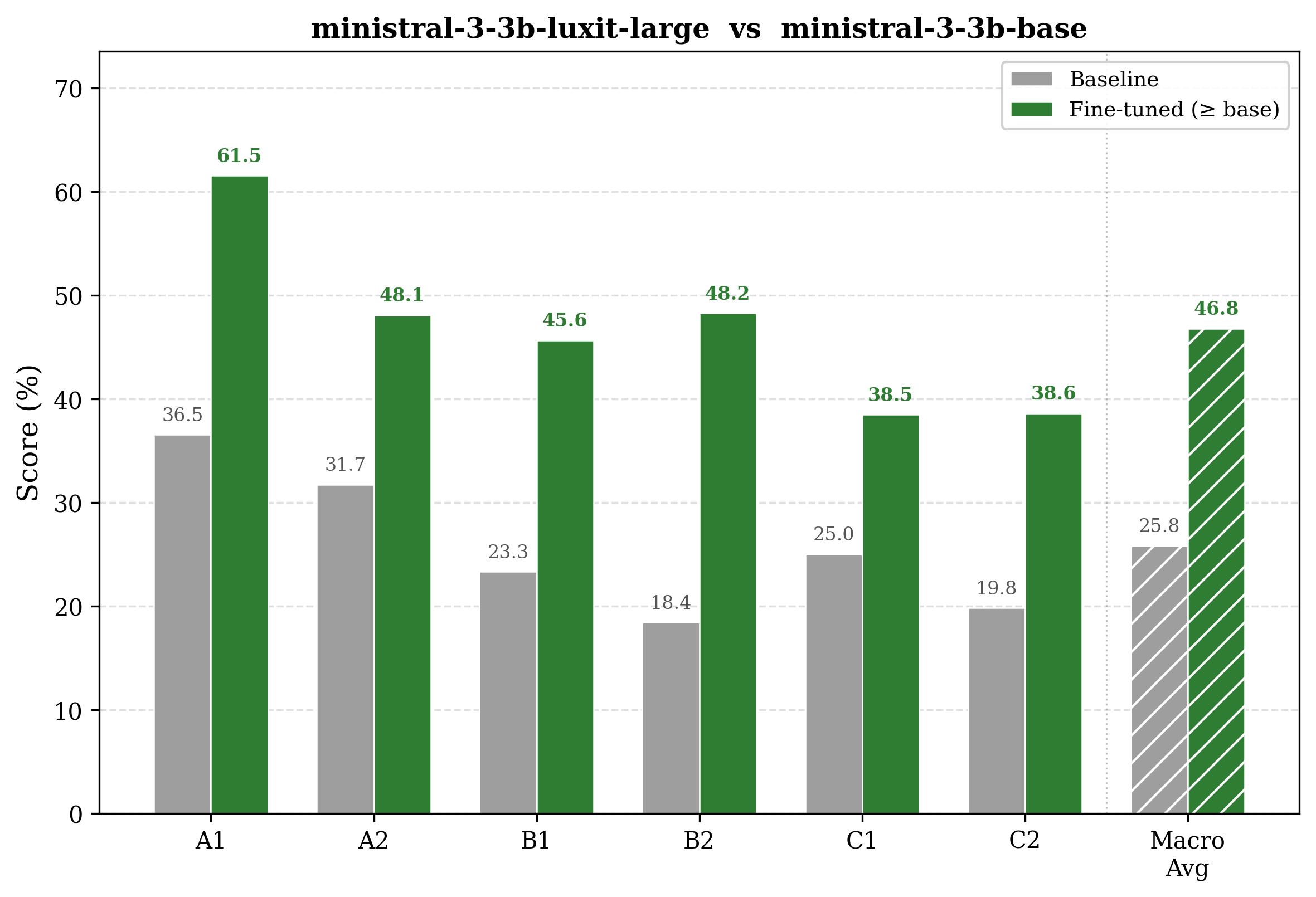}
        \caption{ministral-3-3b}
        \label{fig:ministral-3-3b}
    \end{subfigure}
    \begin{subfigure}[t]{0.45\textwidth}
        \centering
        \includegraphics[width=\textwidth]{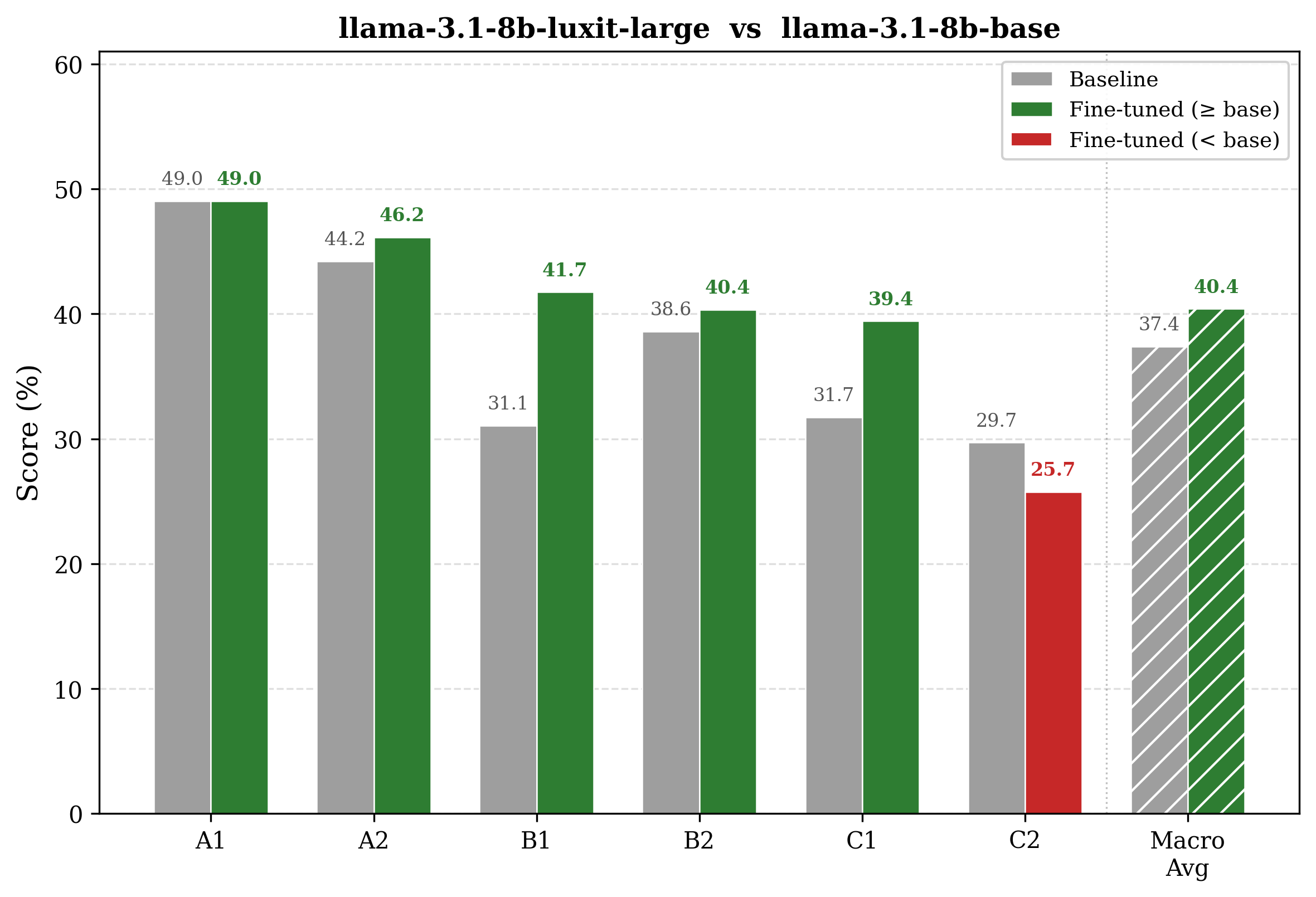}
        \caption{llama-3.1-8b}
        \label{fig:llama-3.1-8b}
    \end{subfigure}
    \hfill
    \begin{subfigure}[t]{0.45\textwidth}
        \centering
        \includegraphics[width=\textwidth]{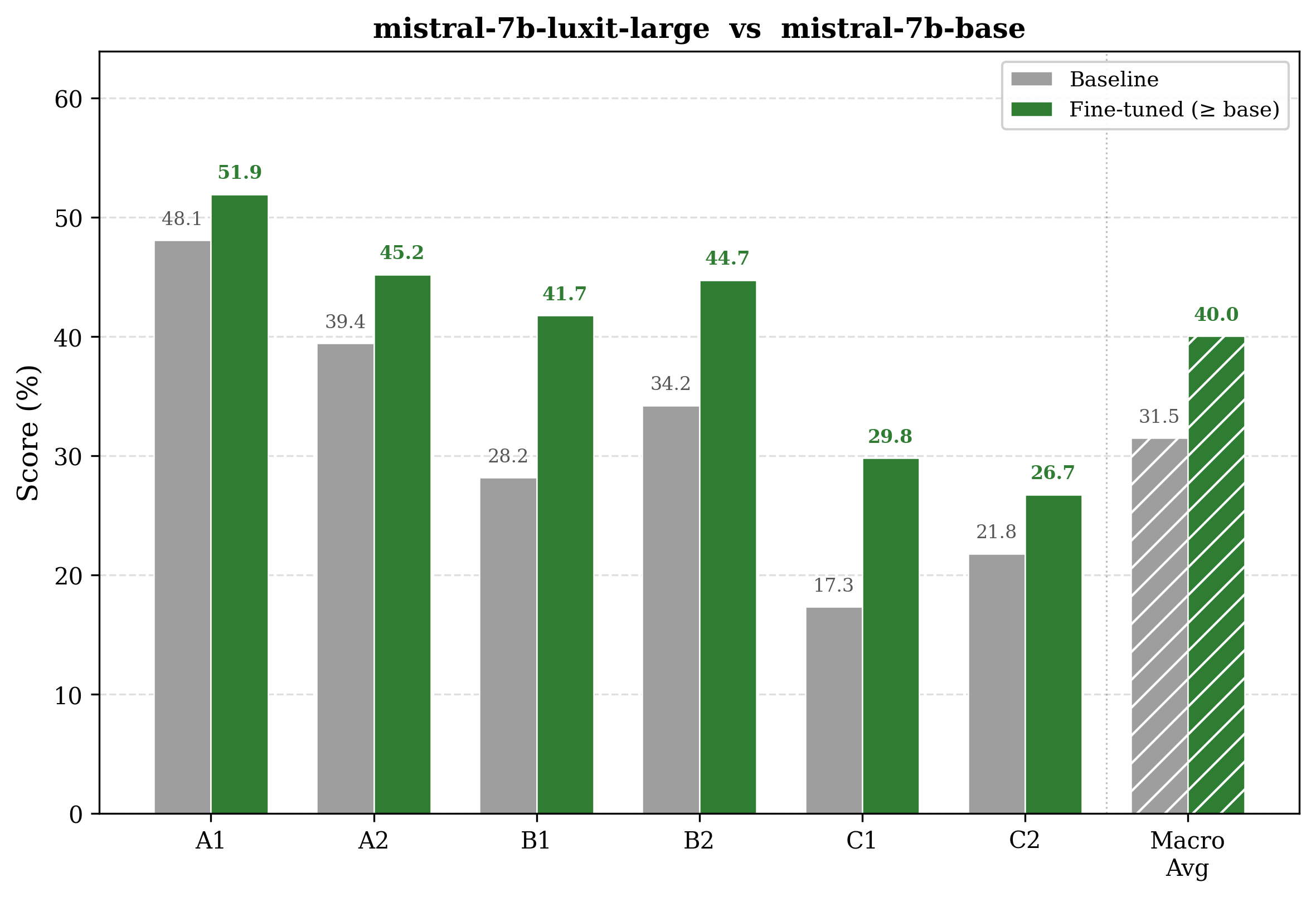}
        \caption{mistral-7b}
        \label{fig:mistral-7b}
    \end{subfigure}
    \caption{Total accuracy of fine-tuned models (labeled luxit-large) compared to their baselines on Luxembourgish language exams (A1--C2). Green bars indicate improvement over baseline; red bars indicate regression. Models are ordered roughly by macro-average performance of the fine-tuned variant (best to worst, left-to-right, top-to-bottom).}
    \label{fig:all-model-comparisons}
\end{figure*}

\begin{figure*}[p]
    \ContinuedFloat
    \centering
    \begin{subfigure}[t]{0.45\textwidth}
        \centering
        \includegraphics[width=\textwidth]{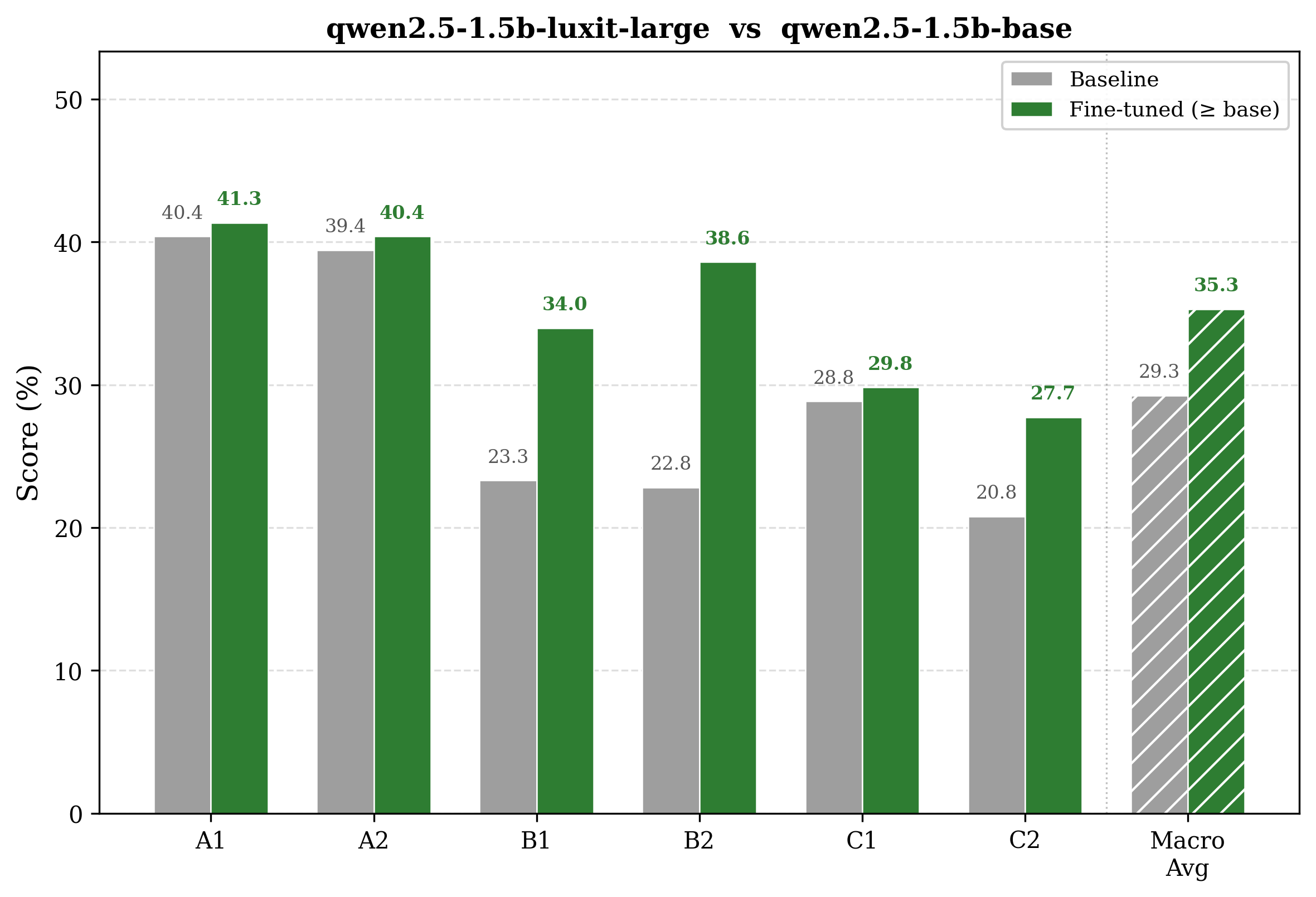}
        \caption{qwen2.5-1.5b}
        \label{fig:qwen2.5-1.5b}
    \end{subfigure}
    \hfill
    \begin{subfigure}[t]{0.45\textwidth}
        \centering
        \includegraphics[width=\textwidth]{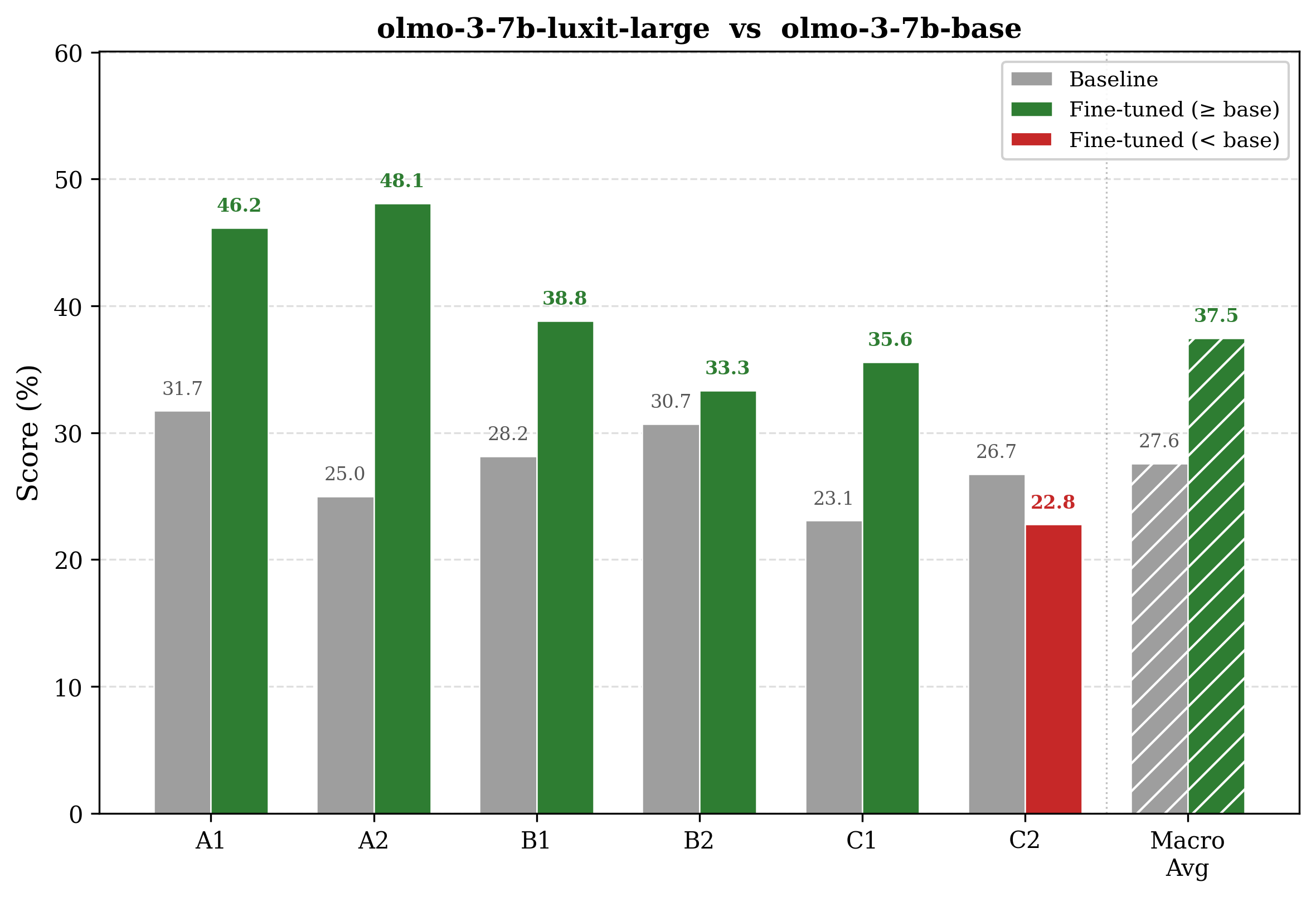}
        \caption{olmo-3-7b}
        \label{fig:olmo-3-7b}
    \end{subfigure}
    \begin{subfigure}[t]{0.45\textwidth}
        \centering
        \includegraphics[width=\textwidth]{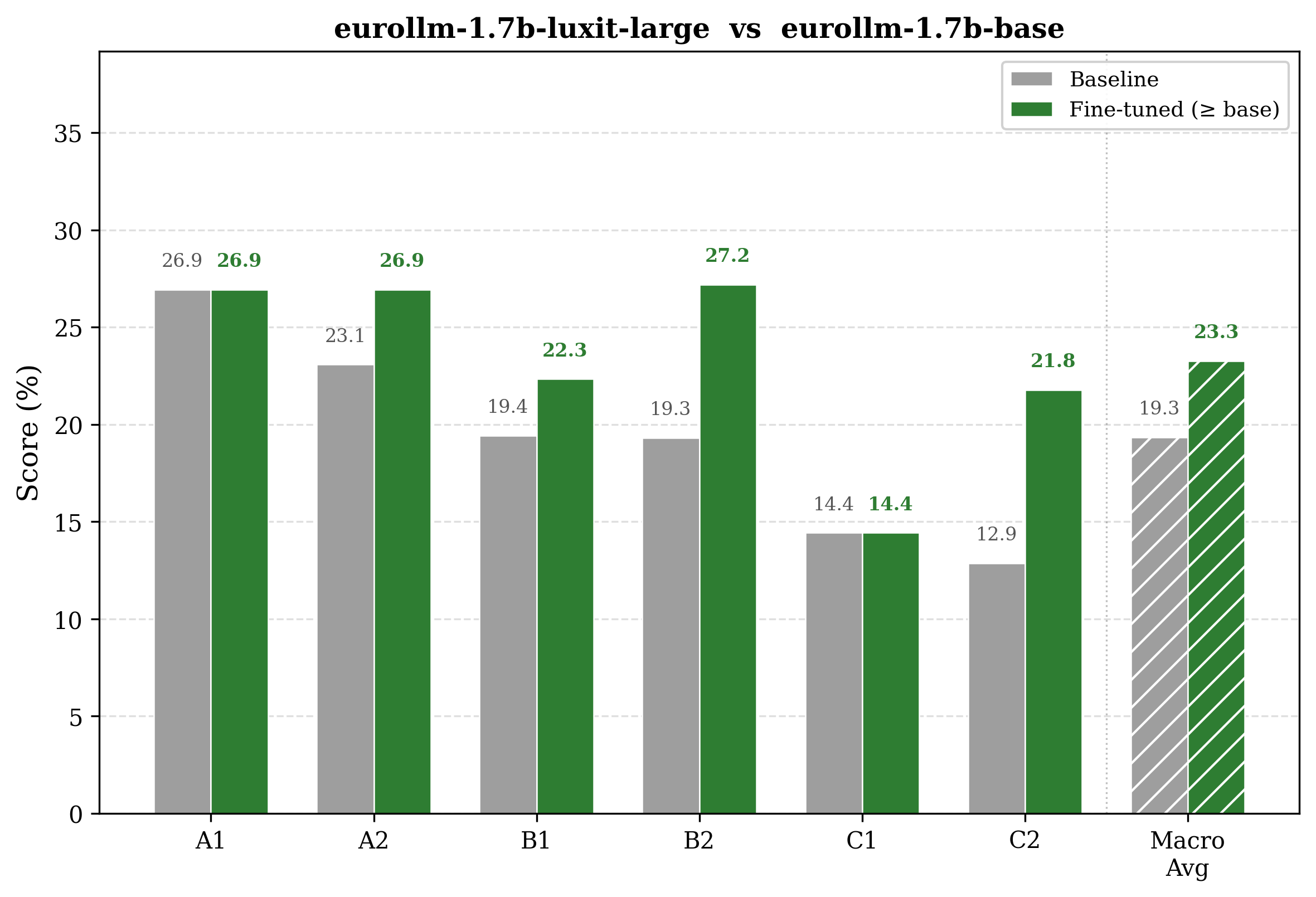}
        \caption{eurollm-1.7b}
        \label{fig:eurollm-1.7b}
    \end{subfigure}
    \hfill
    \begin{subfigure}[t]{0.45\textwidth}
        \centering
        \includegraphics[width=\textwidth]{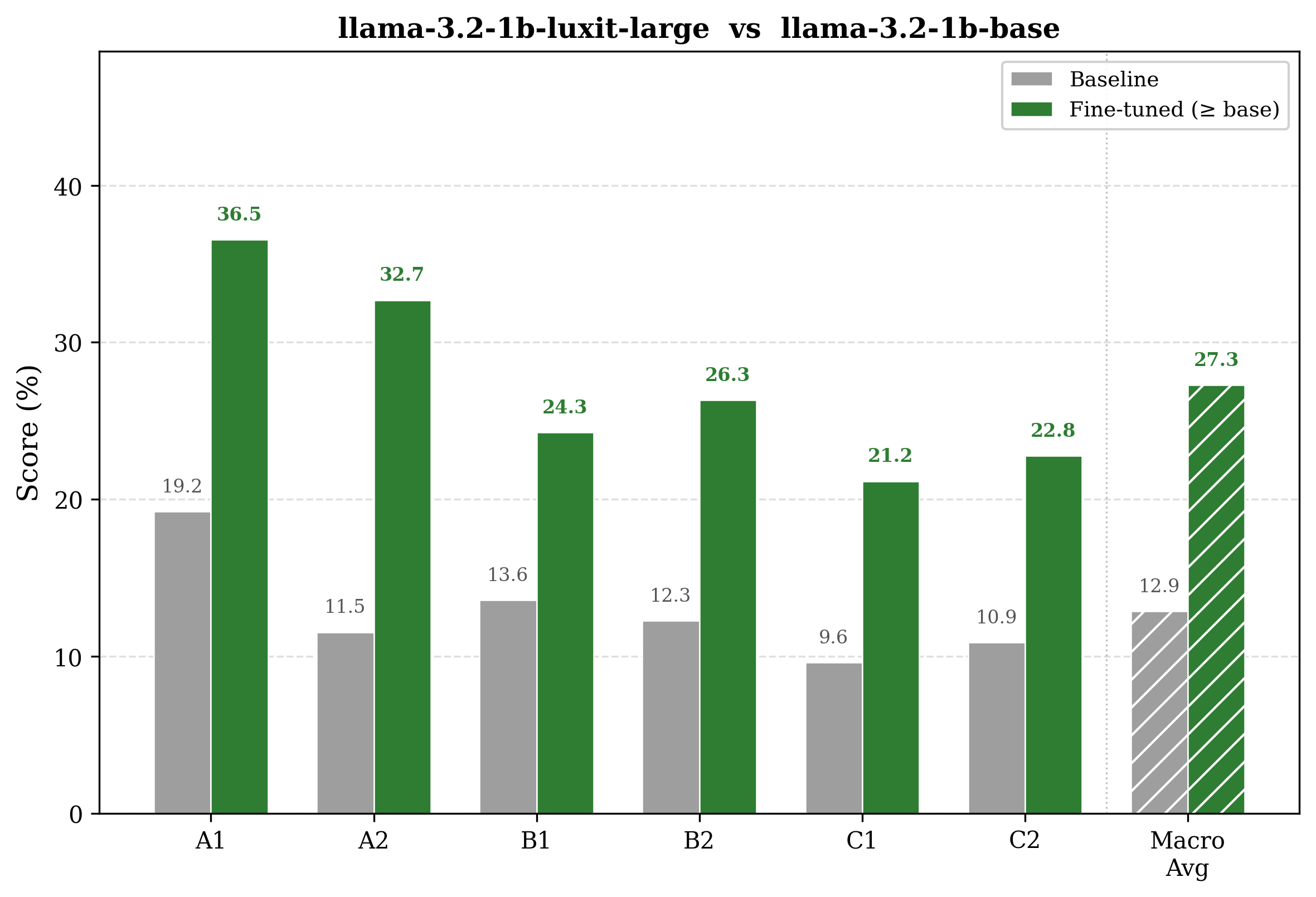}
        \caption{llama-3.2-1b}
        \label{fig:llama-3.2-1b}
    \end{subfigure}
    \begin{subfigure}[t]{0.45\textwidth}
        \centering
        \includegraphics[width=\textwidth]{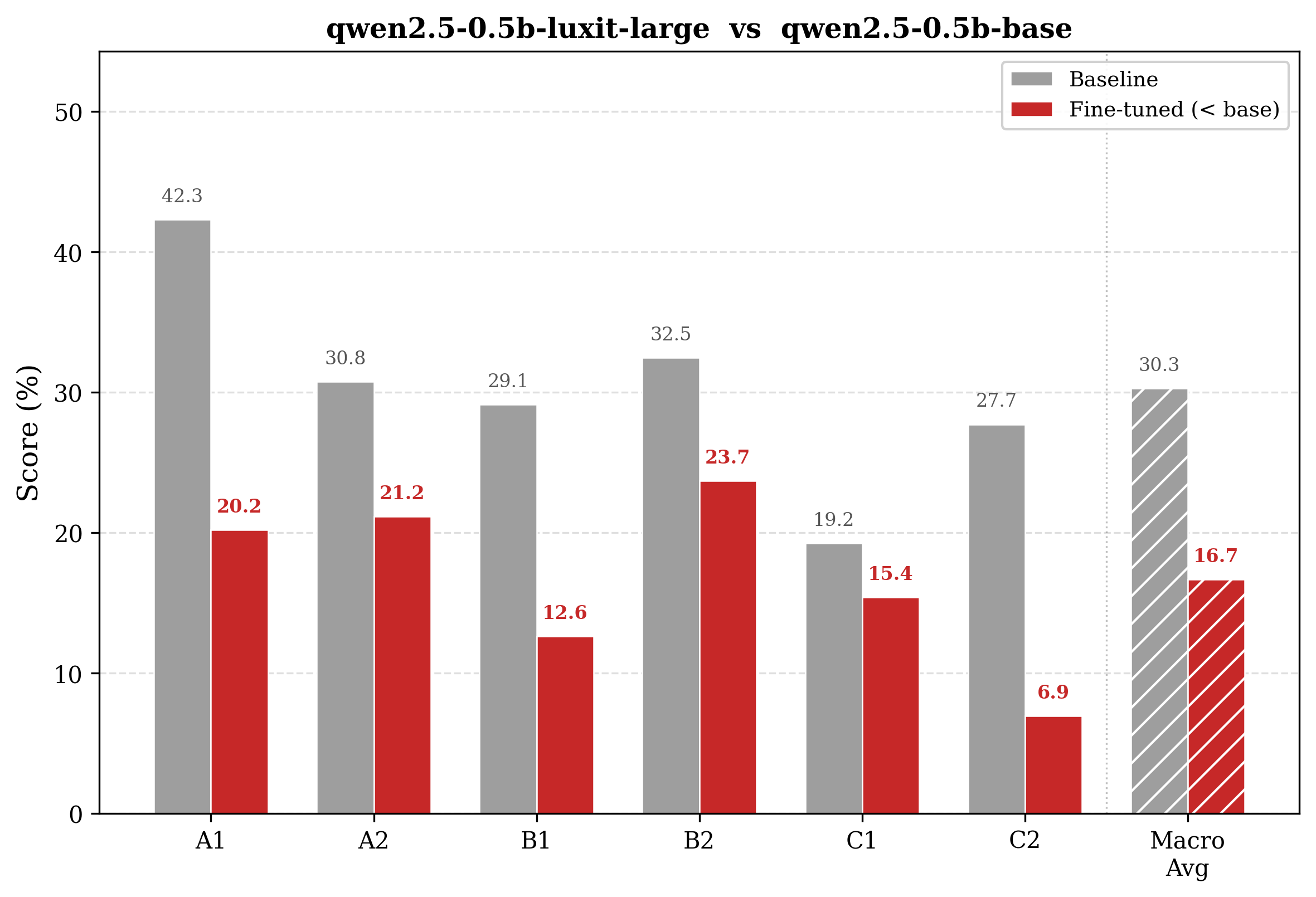}
        \caption{qwen2.5-0.5b}
        \label{fig:qwen2.5-0.5b}
    \end{subfigure}
    \hfill
    \begin{subfigure}[t]{0.45\textwidth}
        \centering
        \includegraphics[width=\textwidth]{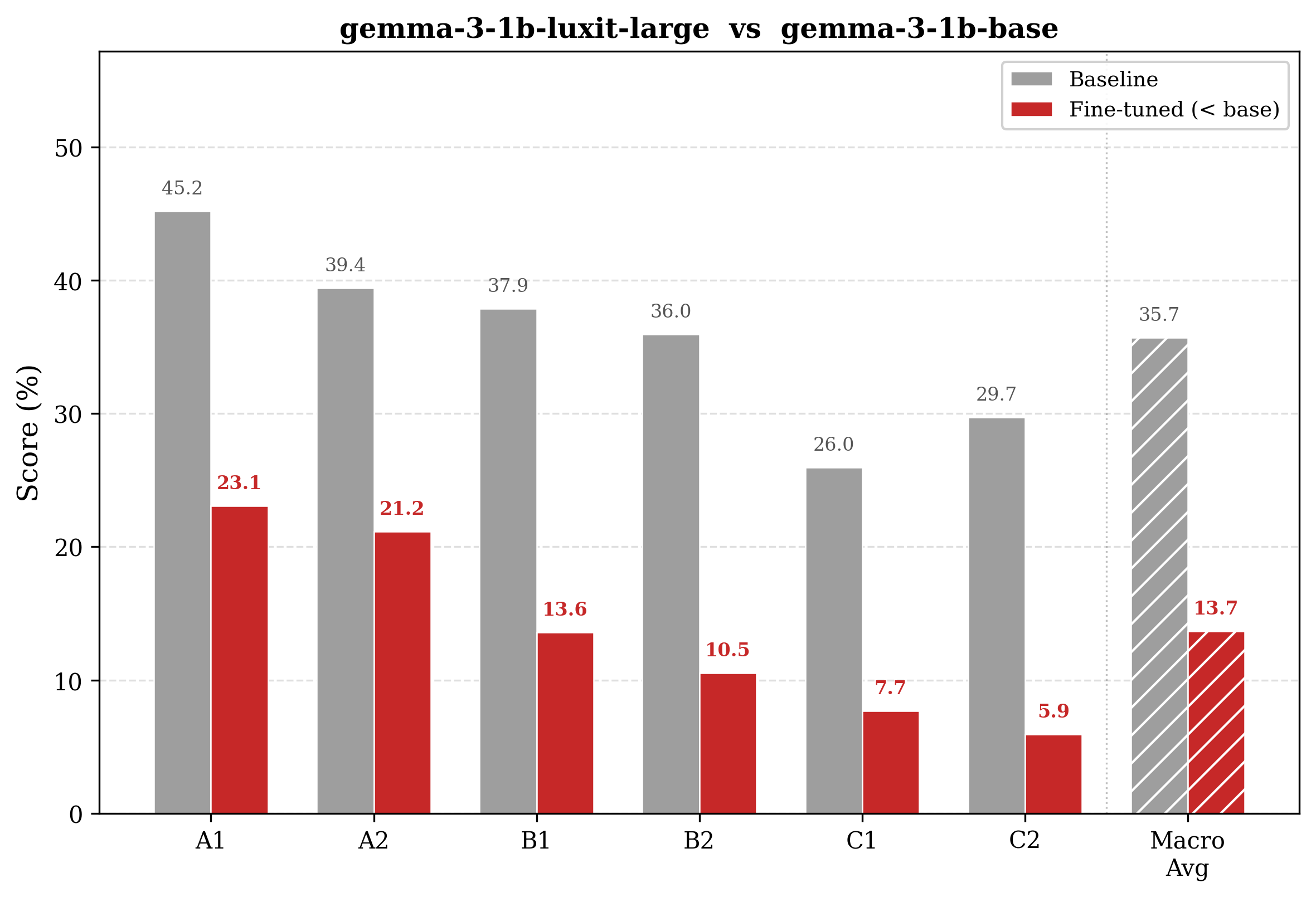}
        \caption{gemma-3-1b}
        \label{fig:gemma-3-1b}
    \end{subfigure}
    \caption[]{(Continued) Total accuracy of fine-tuned models compared to their baselines on Luxembourgish language exams (A1--C2).}
\end{figure*}

\subsection{RQ2}
\label{app:rq3}
\begin{figure*}[ht]
    \centering
    \includegraphics[width=\textwidth]{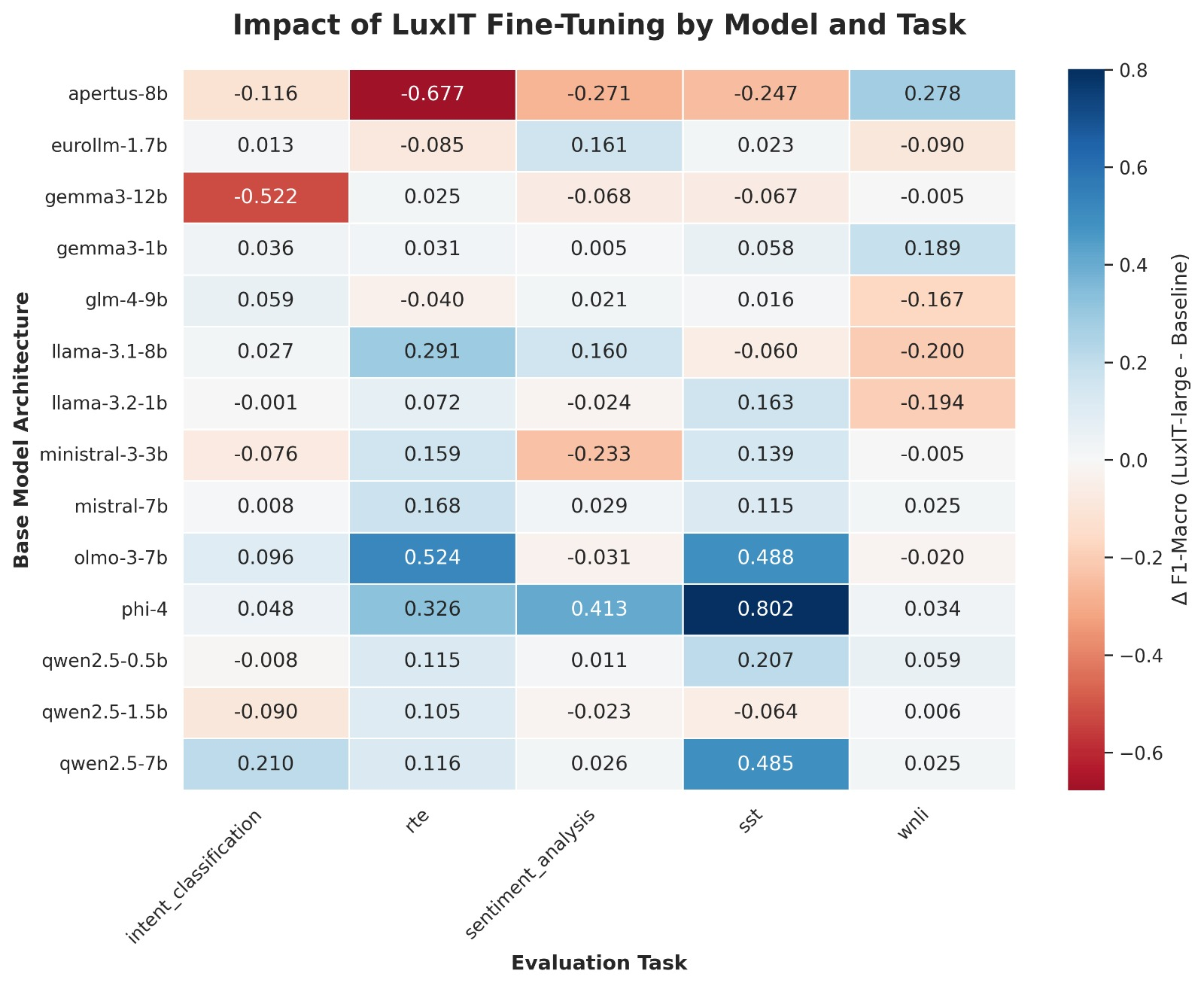}
    \caption{Impact of LuxIT Fine-Tuning by Model and Task ($\Delta$F1-Macro).}
    \label{fig:heatmap_downstream_app}
\end{figure*}

We report the results of benchmarking our fine-tuned models against their respective baseline models (Section \ref{sec: rq3}) on the Luxembourgish downstream tasks in Table \ref{tab:dowstream_task_results} (introduced in Section \ref{sec: eval_downstream_tasks}) and show the per-task changes and the average change per model in Figure \ref{fig:heatmap_downstream_app} and Figure \ref{fig:barplot_downstream} respectively.

\begin{table*}[htbp]
\centering
\resizebox{\textwidth}{!}{%
\renewcommand{\arraystretch}{1.15}
\setlength{\tabcolsep}{5pt}
\begin{tabular}{l|rrr|rrr|rrr|rrr|rrr|r}
\toprule
& \multicolumn{3}{c|}{\textbf{IC}}
& \multicolumn{3}{c|}{\textbf{RTE}}
& \multicolumn{3}{c|}{\textbf{SA}}
& \multicolumn{3}{c|}{\textbf{SST-2}}
& \multicolumn{3}{c|}{\textbf{WNLI}}
& \\
\cmidrule(lr){2-4}\cmidrule(lr){5-7}\cmidrule(lr){8-10}
\cmidrule(lr){11-13}\cmidrule(lr){14-16}
\textbf{Base Model}
  & Base & LuxIT & $\Delta$
  & Base & LuxIT & $\Delta$
  & Base & LuxIT & $\Delta$
  & Base & LuxIT & $\Delta$
  & Base & LuxIT & $\Delta$
  & \textbf{Avg.\ $\Delta$} \\
\midrule
phi-4
  & 0.647 & 0.695 & $+$0.048
  & 0.449 & 0.775 & $+$0.326
  & 0.017 & 0.430 & $+$0.413
  & 0.015 & 0.816 & $+$0.802
  & 0.293 & 0.327 & $+$0.034
  & $\mathbf{+0.325}$ \\
olmo-3-7b
  & 0.150 & 0.245 & $+$0.096
  & 0.142 & 0.666 & $+$0.524
  & 0.249 & 0.218 & $-$0.031
  & 0.300 & 0.787 & $+$0.488
  & 0.285 & 0.265 & $-$0.020
  & $\mathbf{+0.211}$ \\
qwen2.5-7b
  & 0.491 & 0.702 & $+$0.210
  & 0.658 & 0.774 & $+$0.116
  & 0.323 & 0.349 & $+$0.026
  & 0.285 & 0.770 & $+$0.485
  & 0.511 & 0.536 & $+$0.025
  & $\mathbf{+0.172}$ \\
qwen2.5-0.5b
  & 0.008 & 0.000 & $-$0.008
  & 0.276 & 0.392 & $+$0.115
  & 0.065 & 0.076 & $+$0.011
  & 0.297 & 0.504 & $+$0.207
  & 0.331 & 0.390 & $+$0.059
  & $\mathbf{+0.077}$ \\
mistral-7b
  & 0.412 & 0.419 & $+$0.008
  & 0.548 & 0.716 & $+$0.168
  & 0.354 & 0.383 & $+$0.029
  & 0.652 & 0.767 & $+$0.115
  & 0.265 & 0.290 & $+$0.025
  & $\mathbf{+0.069}$ \\
gemma-3-1b
  & 0.150 & 0.186 & $+$0.036
  & 0.530 & 0.560 & $+$0.031
  & 0.246 & 0.251 & $+$0.005
  & 0.636 & 0.694 & $+$0.058
  & 0.293 & 0.482 & $+$0.189
  & $\mathbf{+0.064}$ \\
llama-3.1-8b
  & 0.518 & 0.545 & $+$0.027
  & 0.332 & 0.623 & $+$0.291
  & 0.188 & 0.348 & $+$0.160
  & 0.711 & 0.651 & $-$0.060
  & 0.520 & 0.320 & $-$0.200
  & $\mathbf{+0.044}$ \\
eurollm-1.7b
  & 0.006 & 0.019 & $+$0.013
  & 0.139 & 0.054 & $-$0.085
  & 0.052 & 0.213 & $+$0.161
  & 0.267 & 0.289 & $+$0.023
  & 0.318 & 0.227 & $-$0.090
  & $\mathbf{+0.004}$ \\
llama-3.2-1b
  & 0.010 & 0.009 & $-$0.001
  & 0.251 & 0.324 & $+$0.072
  & 0.106 & 0.082 & $-$0.024
  & 0.175 & 0.338 & $+$0.163
  & 0.459 & 0.265 & $-$0.194
  & $\mathbf{+0.003}$ \\
\midrule
ministral-3-3b
  & 0.398 & 0.322 & $-$0.076
  & 0.392 & 0.551 & $+$0.159
  & 0.425 & 0.192 & $-$0.233
  & 0.545 & 0.684 & $+$0.139
  & 0.329 & 0.324 & $-$0.005
  & $-$0.003 \\
qwen2.5-1.5b
  & 0.232 & 0.142 & $-$0.090
  & 0.310 & 0.414 & $+$0.105
  & 0.263 & 0.240 & $-$0.023
  & 0.565 & 0.501 & $-$0.064
  & 0.385 & 0.390 & $+$0.006
  & $-$0.013 \\
glm-4-9b
  & 0.684 & 0.743 & $+$0.059
  & 0.775 & 0.735 & $-$0.040
  & 0.307 & 0.328 & $+$0.021
  & 0.823 & 0.838 & $+$0.016
  & 0.478 & 0.311 & $-$0.167
  & $-$0.022 \\
gemma-3-12b
  & 0.865 & 0.342 & $-$0.522
  & 0.744 & 0.770 & $+$0.025
  & 0.524 & 0.455 & $-$0.068
  & 0.783 & 0.716 & $-$0.067
  & 0.569 & 0.564 & $-$0.005
  & $-$0.128 \\
apertus-8b
  & 0.576 & 0.460 & $-$0.116
  & 0.686 & 0.010 & $-$0.677
  & 0.505 & 0.234 & $-$0.271
  & 0.706 & 0.459 & $-$0.247
  & 0.265 & 0.543 & $+$0.278
  & $-$0.206 \\
\bottomrule
\end{tabular}%
}
\caption{%
  Complete evaluation results comparing \textbf{Baseline} and \textbf{LuxIT} fine-tuned
  F1-Macro scores across five Luxembourgish NLP tasks:
  Intent Classification (IC), Recognizing Textual Entailment (RTE),
  Sentiment Analysis (SA), Stanford Sentiment Treebank (SST-2), and
  Winograd Natural Language Inference (WNLI).
  $\Delta$ = LuxIT $-$ Baseline; bold values in the
  \emph{Avg.\ $\Delta$} column indicate net improvement.
  Models are sorted by average $\Delta$ in descending order;
  the horizontal rule separates positive from negative average gains.%
}
\label{tab:dowstream_task_results}
\end{table*}

\end{document}